\documentclass[a4paper,12pt]{article}

\usepackage{float}
\usepackage{amsmath} 
\usepackage{amssymb}
\usepackage{amsfonts}
\usepackage{epsfig}
\usepackage{color,fancybox,graphicx}

\numberwithin{equation}{section}

\newtheorem{lem}{Lemma}[section]
\newtheorem{cor}{Corollary}[section]
\newtheorem{pro}{Proposition}[section]
\newtheorem{theo}{Theorem}[section]
\newtheorem{fact}{Fact}[section]

\newcommand{\bX}{\mathbf{X}}
\newcommand{\bx}{\mathbf{x}}
\newcommand{\lk}{\lceil{\log_2 k_n}\rceil}

\begin{document}

\begin{center}

{\sc \Large Analysis of a Random Forests Model\\

\vspace{0.5cm}}

G\'erard Biau
\vspace{0.2cm}\\
LSTA \& LPMA\footnote{Research partially supported by the French National Research Agency under grant ANR-09-BLAN-0051-02 ``CLARA''.}\\
Universit\'e Pierre et Marie Curie -- Paris VI\\
Bo\^{\i}te 158,  Tour 15-25, 2\`eme \'etage\\
4 place Jussieu, 75252 Paris Cedex 05, France

\medskip
---
\medskip

DMA\footnote{Research carried out within the INRIA project ``CLASSIC'' hosted by Ecole Normale Sup{\'e}rieure and CNRS.}\\
Ecole Normale Sup\'erieure\\
45 rue d'Ulm\\
75230 Paris Cedex 05, France\\
\smallskip
\texttt{gerard.biau@upmc.fr}
\vspace{0.5cm}

\end{center}
\begin{abstract} 
\noindent {\rm Random forests are a scheme proposed by Leo Breiman in the 2000's for building a predictor ensemble with a set of decision trees that grow in randomly selected subspaces of data. Despite growing interest and  practical use, there has been little exploration of the statistical properties of random forests, and little is known about the mathematical forces driving the algorithm.  In this paper, we offer an in-depth analysis of a random forests model suggested by Breiman in \cite{Bre04}, which is very close to the original algorithm. We show in particular that the procedure is consistent and adapts to sparsity, in the sense that its rate of convergence depends only on the number of strong features and not on how many noise variables are present.
\medskip

\medskip
 
}

\end{abstract}

\setcounter{footnote}{0}
\section{Introduction}
\subsection{Random forests}
In a series of papers and technical reports, Breiman \cite{Bre96,Bre00,Bre01,Bre04}  demonstrated that substantial gains in classification and regression accuracy can be achieved by using ensembles of trees, where each tree in the ensemble is grown in accordance with  a random parameter. Final predictions are obtained by aggregating over the ensemble. As the base constituents of the ensemble are tree-structured predictors, and since each of these trees is constructed using an injection of randomness, these procedures are called ``random forests''. 
\medskip

Breiman's ideas were decisively influenced by the early work of Amit and Geman \cite{Amit} on geometric feature selection, the random subspace method of Ho \cite{Ho} and the random split selection approach of Dietterich \cite{D2000}. As highlighted by various empirical studies (see \cite{Bre01, Svetnik, Diaz, Genuer, Genuer2} for instance), random forests have emerged as serious competitors to state-of-the-art methods such as boosting (Freund \cite{Freund}) and support vector machines (Shawe-Taylor and Cristianini \cite{STC}). They are fast and easy to implement, produce highly accurate predictions and can handle a very large number of input variables without overfitting. In fact, they are considered to be one of the most accurate general-purpose learning techniques available. The survey by Genuer et al.~\cite{Genuer} may provide the reader with practical guidelines and a good starting point for understanding  the method.
\medskip

In Breiman's approach, each tree in the collection is formed by first selecting at random, at each node, a small group of input coordinates (also called features or variables hereafter) to split on and, secondly, by calculating the best split based on these features in the training set. The tree is grown using CART methodology (Breiman et al.~\cite{Breimanbook}) to maximum size, without pruning. This subspace randomization scheme is blended with bagging (\cite{Bre96, BY, Bplus, BCG})  to resample, with replacement, the training data set each time a new individual tree is grown. 
\medskip

Although the mechanism appears simple, it involves many different driving forces which make it difficult to analyse. In fact, its mathematical properties remain to date largely unknown and, up to now, most theoretical studies have concentrated on isolated parts or stylized versions of the algorithm.  Interesting attempts in this direction are by Lin and Jeon \cite{Lin}, who establish a connection between random forests
and adaptive nearest neighbor methods (see also \cite{Zepaper} for further results); Meinshausen \cite{M}, who studies the consistency of random forests in the context of conditional quantile prediction; and Devroye et al.~\cite{BDL08}, who offer consistency theorems for various simplified versions
of random forests and other randomized ensemble predictors. Nevertheless, the statistical mechanism
of ``true'' random forests is not yet fully understood and is still under active investigation.
\medskip

In the present paper, we go one step further into  random forests by working out and solidifying the properties of a model suggested by Breiman in \cite{Bre04}. Though this model is still simple compared to the ``true'' algorithm, it is nevertheless closer to reality than any other scheme we are aware of. The short draft \cite{Bre04} is essentially based on intuition and mathematical heuristics,  some of them are questionable and make the document difficult to read and understand. However, the ideas presented by Breiman are worth clarifying and developing, and they will serve as a starting point for our study.
\medskip

Before we formalize the model, some definitions are in order. Throughout the document, we suppose that we are given a training sample $\mathcal D_n=\{(\bX_1,Y_1), \hdots, (\bX_n,Y_n)\}$ of i.i.d. $[0,1]^d\times \mathbb R$-valued random variables ($d \geq 2$) with the same distribution as an independent generic pair $(\bX, Y)$ satisfying $\mathbb E Y^2 < \infty$. The space $\mathbb [0,1]^d$ is equipped with the standard Euclidean metric. For  fixed $\bx \in [0,1]^d$, our goal is to estimate the regression function $r(\bx)=\mathbb E[Y|\bX=\bx]$ using the data $\mathcal D_n$. In this respect, we say that a regression function estimate $r_n$ is consistent if $\mathbb E [r_n(\bX)-r(\bX)]^2 \to 0$ as $n \to \infty$. The main message of this paper is that Breiman's procedure is consistent and adapts to sparsity, in the sense that its rate of convergence depends only on the number of strong features and not on how many noise variables are present.
\subsection{The model}
Formally, a random forest is a predictor consisting of a collection of randomized base regression trees $\{r_n(\bx, \Theta_m,\mathcal D_n), m\geq 1\}$, where $\Theta_1, \Theta_2,\hdots$ are i.i.d. outputs of a randomizing variable $\Theta$. These random trees are combined to form the aggregated regression estimate
$$\bar r_n(\bX, \mathcal D_n)=\mathbb E_{\Theta}\left[r_n(\bX,\Theta, \mathcal D_n)\right],$$
where $\mathbb E_{\Theta}$ denotes expectation with respect to the random parameter, conditionally on $\bX$ and the data set $\mathcal D_n$.  In the following, to lighten notation a little, we will omit the dependency of the estimates in the sample, and write for example $\bar r_n(\bX)$ instead of $\bar r_n(\bX,\mathcal D_n)$. Note that, in practice, the above expectation is evaluated by Monte Carlo, i.e., by generating $M$ (usually large) random trees, and taking the average of the individual outcomes (this procedure is justified by the law of large numbers, see the appendix in Breiman \cite{Bre01}). The randomizing variable $\Theta$ is used to determine how the successive cuts are performed when building the individual trees, such as selection of the coordinate to split and position of the split. \medskip

In the model we have in mind, the variable $\Theta$ is assumed to be independent of $\bX$ and the training sample $\mathcal D_n$. This excludes in particular any bootstrapping or resampling step in the training set. This also rules out any data-dependent strategy to build the trees, such as searching for optimal splits by optimizing some criterion on the actual observations. However, we allow $\Theta$  to be based on a second sample, independent of, but distributed as, $\mathcal D_n$. This important issue will be thoroughly discussed in Section 3. 
\medskip

With these warnings in mind, we will assume that each individual random tree is constructed in the following way. All nodes of the tree are associated with rectangular
cells such that at each step of the construction of the tree, the
collection of cells associated with the leaves of the tree (i.e., 
external nodes) forms a partition of $[0,1]^d$. The root of the tree is $[0,1]^d$ itself. The following  procedure is then repeated $\lk$ times, where $\log_2$ is the base-2 logarithm, $\lceil . \rceil$ the ceiling function and $k_n\ge 2$ a deterministic
parameter, fixed beforehand by the user, and possibly depending on $n$. 
\begin{enumerate}
\item At each node, a coordinate of $\bX=(X^{(1)}, \hdots, X^{(d)})$ is selected, with the $j$-th feature having a probability $p_{nj} \in (0,1)$ of being selected.
\item At each node, once the coordinate is selected, the split is at the midpoint of the chosen side.
\end{enumerate}
Each randomized tree $r_n(\bX,\Theta)$ outputs the average over all $Y_i$ for which the corresponding vectors $\bX_i$ fall in the same cell of the random partition as $\bX$. In other words, letting $A_n(\bX,\Theta)$ be the rectangular cell of the random partition
containing $\bX$,
$$r_n(\bX,\Theta)=\frac{\sum_{i=1}^n Y_i \mathbf 1_{[\bX_i \in A_n(\bX,\Theta)]}}{\sum_{i=1}^n \mathbf 1_{[\bX_i \in A_n(\bX,\Theta)]}}\,\mathbf 1_{\mathcal E_n(\bX,\Theta)},$$
where the event $\mathcal E_n(\bX,\Theta)$ is defined by
$$\mathcal E_n(\bX,\Theta)=\left [\sum_{i=1}^n \mathbf 1_{[\bX_i \in A_n(\bX,\Theta)]} \neq 0 \right ].$$
(Thus, by convention, the estimate is set to $0$ on empty cells.) Taking finally expectation with respect to the parameter $\Theta$,  the random forests regression estimate takes the form
$$\bar r_n(\bX)=\mathbb E_{\Theta} \left [r_n(\bX,\Theta)\right]= \mathbb E_{\Theta} \left [\frac{\sum_{i=1}^n Y_i \mathbf 1_{[\bX_i \in A_n(\bX,\Theta)]}}{\sum_{i=1}^n \mathbf 1_{[\bX_i \in A_n(\bX,\Theta)]}}\,\mathbf 1_{\mathcal E_n(\bX,\Theta)} \right].$$
Let us now make some general remarks about this random forests model. First of all, we note that, by construction,  each individual tree has exactly $2^{\lk}$ ($\approx k_n$) terminal nodes, and each leaf has Lebesgue measure $2^{-\lk}$ ($\approx 1/k_n$). Thus, if $\bX$ has uniform distribution on $[0,1]^d$, there will be on average about $n/k_n$ observations per terminal node. In particular, the choice $k_n=n$ induces a very small number of cases in the final leaves, in accordance with the idea that the single trees should not be pruned. 
\medskip

Next, we see that, during the construction of the tree, at each node, each candidate coordinate $X^{(j)}$ may be chosen  with probability $p_{nj} \in (0,1)$. This implies in particular $\sum_{j=1}^d p_{nj}=1$. Although we do not precise for the moment the way these probabilities are generated, we stress that they may be induced by a second sample. This includes the situation where, at each node, randomness is introduced by selecting at random (with or without replacement) a small group of input features to split on, and choosing  to cut the cell along the coordinate---inside this group---which most decreases some empirical criterion evaluated on the extra sample. This scheme is close to what the original random forests algorithm does, the essential difference being that the latter algorithm uses the actual data set to calculate the best splits. This point will be properly discussed in Section 3.
\medskip

Finally, the requirement that the splits are always achieved at the middle of the cell sides is mainly technical, and it could eventually be replaced by a more involved random mechanism---based on the second sample---, at the price of a much more complicated analysis.
\medskip

The document is organized as follows. In Section 2, we prove that the random forests regression estimate $\bar r_n$ is consistent and discuss its rate of convergence. As a striking result, we show under a sparsity framework that the rate of convergence depends only on the number of active (or strong) variables and not on the dimension of the ambient space. This feature is particularly desirable in high-dimensional  regression, when the number of variables
can be much larger than the sample size, and may explain why random forests are able to handle a very large number of input variables without overfitting. Section 3 is devoted to a discussion, and a small simulation study is presented in Section 4. For the sake of clarity, proofs are postponed to Section 5.
\section{Asymptotic analysis}
Throughout the document, we denote by $N_n(\bX,\Theta)$ the number of data points falling in the same cell as $\bX$, i.e.,
$$
N_n(\bX,\Theta)= \sum_{i=1}^n \mathbf 1_{[\bX_i\in A_n(\bX,\Theta)]}.$$
We start the analysis with the following simple theorem, which shows that the random forests estimate $\bar r_n$ is consistent.
\begin{theo}
\label{convergence}
Assume that the distribution of $\bX$ has support on $[0,1]^d$. Then the random forests estimate $\bar{r}_n$ is consistent
whenever $p_{nj}\log k_n\to \infty$ for all $j=1, \hdots,d$ and $k_n/n \to 0$ as $n\to \infty$.
\end{theo}

Theorem \ref{convergence} mainly serves as an illustration of how the consistency problem of random forests predictors may be attacked. It encompasses, in particular, the situation where, at each node, the coordinate to split is chosen uniformly at random over the $d$ candidates. In this ``purely random'' model, $p_{nj}=1/d$, independently of $n$ and $j$, and consistency is ensured as long as $k_n\to \infty$ and $k_n/n \to 0$. This is however a radically simplified version of the random forests used in practice, which does not explain the good performance of the algorithm. To achieve this goal, a more in-depth analysis is needed. 
\medskip

There is empirical evidence that many signals in high-dimensional spaces admit a
sparse representation.  As an example, wavelet coefficients of images often exhibit
exponential decay, and a relatively small subset of all wavelet coefficients allows for a good approximation of the original image. Such signals have few non-zero coefficients
and can therefore be described as sparse in the signal
domain (see for instance \cite{Bruckstein}). Similarly, recent advances in high-throughput technologies---such as array comparative genomic hybridization---indicate that, despite the huge dimensionality of problems, only a small number of genes may play a role in determining the outcome and be required to create good predictors (\cite{Veer} for instance). Sparse estimation is playing an increasingly important role in the statistics and machine learning communities, and several methods have recently been developed in both fields, which rely upon the notion of sparsity (e.g.~penalty methods like the Lasso and Dantzig selector, see \cite{Tibshirani,  Candes, Bunea, Bickel} and the references therein). 
\medskip

Following this idea, we will assume in our setting that the target regression function $r(\bX)=\mathbb E[Y|\bX]$, which is initially a function of $\bX=(X^{(1)}, \hdots, X^{(d)})$, depends in fact only on a nonempty subset $\mathcal S$ (for $\mathcal S$trong) of the $d$ features. In other words, letting $\bX_{\mathcal S}=(X_j\,:\,j \in \mathcal S)$ and $S=\mbox{Card } \mathcal S$, we have
$$r(\bX)=\mathbb E[Y|\bX_{\mathcal S}]$$
or equivalently, for any $\bx \in \mathbb [0,1]^d$,
\begin{equation}
\label{sparse}
r(\bx)=r^{\star}(\bx_{\mathcal S}) \quad \mu\mbox{-a.s.},
\end{equation}
where $\mu$ is the distribution of $\bX$ and $r^{\star}:[0,1]^S\to \mathbb R$ is the section of $r$ corresponding to $\mathcal S$. To avoid trivialities, we will assume throughout that $\mathcal S$ is nonempty, with $S\geq 2$. The variables in the set $\mathcal W=\{1, \hdots, d\}-\mathcal S$ (for $\mathcal W$eak) have thus no influence on the response and could be safely removed. In the dimension reduction scenario we have in mind, the ambient dimension $d$ can be very large, much larger than the sample size $n$, but we believe that the representation is sparse, i.e., that very few coordinates of $r$ are non-zero, with indices corresponding to the set $\mathcal S$. Note however that representation (\ref{sparse}) does not forbid the somehow undesirable case where $S=d$. As such, the value $S$ characterizes the sparsity of the model: The smaller $S$, the sparser $r$.
\medskip

Within this sparsity framework, it is intuitively clear that the coordinate-sampling probabilities should ideally satisfy the constraints $p_{nj}=1/S$ for $j \in \mathcal S$ (and, consequently, $p_{nj}=0$ otherwise). However, this is a too strong requirement, which has no chance to be satisfied in practice, except maybe in some special situations where we know beforehand which variables are important and which are not. Thus, to stick to reality, we will rather require in the following that $p_{nj}=(1/S)(1+{\xi}_{nj})$ for $j \in \mathcal S$ (and $p_{nj}={\xi}_{nj}$ otherwise), where $p_{nj} \in (0,1)$ and each ${\xi}_{nj}$ tends to $0$ as $n$ tends to infinity. We will see in Section 3 how to design a randomization mechanism to obtain such probabilities, on the basis of a second sample independent of the training set $\mathcal D_n$. At this point, it is important to note that the dimensions $d$ and $S$ are held constant throughout the document. In particular, these dimensions are {\it not} functions of the sample size $n$, as it may be the case in other asymptotic studies. 
\medskip

We have now enough material for a deeper understanding of the random forests algorithm. To lighten notation a little, we will write
$$W_{ni}(\bX,\Theta)=\frac{\mathbf 1_{[\bX_i \in A_n(\bX,\Theta)]}}{N_n(\bX,\Theta)}\,\mathbf 1_{\mathcal E_n(\bX,\Theta)},$$
so that the estimate takes the form
$$\bar r_n(\bX)=\sum_{i=1}^n \mathbb E_{\Theta} \left[W_{ni}(\bX,\Theta)\right]Y_i.
$$
Let us start with the variance/bias decomposition
\begin{equation}
\label{bv}
\mathbb E \left[\bar r_n(\bX)- r(\bX) \right]^2=\mathbb E \left[{\bar r_n}(\bX)-\tilde {r}_n(\bX) \right]^2+\mathbb E \left[\tilde r_n(\bX)-r(\bX) \right]^2,
\end{equation}
where we set
$$\tilde r_n(\bX)= \sum_{i=1}^n \mathbb E_{\Theta}\left[W_{ni}(\bX,\Theta)\right]r(\bX_i).$$
The two terms of (\ref{bv}) will be examined separately, in Proposition \ref{variance2} and Proposition \ref{biais}, respectively. Throughout, the symbol $\mathbb V$ denotes variance.
\begin{pro}
\label{variance2}
Assume that $\bX$ is uniformly distributed on $[0,1]^d$ and, for all $\bx \in \mathbb R^d$,
$$\sigma^2(\bx)=\mathbb V[Y\,|\,\bX=\bx]\leq \sigma^2$$
for some positive constant $\sigma^2$. Then, if $p_{nj}=(1/S)(1+{\xi}_{nj})$ for $j \in \mathcal S$,
$$\mathbb E \left [\bar r_n(\bX)-\tilde r_n(\bX) \right]^2  \leq C\sigma^2\left(\frac{S^2}{S-1}\right)^{S/2d}\left(1+\xi_n\right)\frac{k_n}{n(\log k_n)^{S/2d}},$$
where
$$C=\frac{288}{\pi}\left(\frac{\pi \log 2}{16}\right)^{S/2d}.$$
The sequence $(\xi_n)$ depends on the sequences $\{(\xi_{nj}) : j\in \mathcal S\}$ only and tends to $0$ as $n$ tends to infinity.
\end{pro}
{\bf Remark 1}\quad A close inspection of the end of the proof of Proposition \ref{variance2} reveals that
$$1+\xi_n=\prod_{j \in \mathcal S} \left [\left( 1+ \xi_{nj} \right)^{-1} \left (1- \frac{\xi_{nj}}{S-1} \right)^{-1}\right]^{1/2d}.$$
In particular, if $a<p_{nj}<b$ for some constants $a,b \in (0,1)$, then
$$1+\xi_n \leq \left ( \frac{S-1}{S^2a(1-b)}\right)^{S/2d}.$$
\hfill $\blacksquare$
\medskip

The main message of Proposition \ref{variance2} is that the variance of the forests estimate is $\mathcal O(k_n/(n (\log k_n)^{S/2d}))$. This result is interesting by itself since it shows the effect of aggregation on the variance of the forest. To understand this remark, recall that individual (random or not) trees are proved to be consistent by letting the number of cases in each terminal node become large (see \cite[Chapter 20]{DGL}), with a typical variance of the order $k_n/n$. Thus, for such trees, the choice $k_n=n$ (i.e., about one observation on average in each terminal node) is clearly not suitable and leads to serious overfitting and variance explosion. On the other hand, the variance of the forest is of the order $k_n/(n (\log k_n)^{S/2d})$. Therefore, letting $k_n=n$, the variance is of the order $1/(\log n)^{S/2d}$, a quantity which still goes to $0$ as $n$ grows! Proof of Proposition \ref{variance2} reveals that this $\log$ term is a by-product of the $\Theta$-averaging process, which appears by taking into consideration the correlation between trees. We believe that it provides an interesting perspective on why random forests are still able to do a good job, despite the fact that individual trees are not pruned.
\medskip

Note finally that the requirement that $\bX$ is uniformly distributed on the hypercube could be safely replaced by the assumption that $\bX$ has a density with respect to the Lebesgue measure on $[0,1]^d$ and the density is bounded from above and from below. The case where the density of $\bX$ is not bounded from below necessitates a specific analysis, which we believe is beyond the scope of the present paper. We refer the reader to \cite{Zepaper} for results in this direction (see also Remark 5 in Section 5).
\medskip

Let us now turn to the analysis of the bias term in equality (\ref{bv}). Recall that $r^{\star}$ denotes the section of $r$ corresponding to $\mathcal S$.
\begin{pro}
\label{biais}
Assume that  $\bX$ is uniformly distributed on $[0,1]^d$ and $r^{\star}$ is $L$-Lipschitz on $[0,1]^S$. Then, if $p_{nj}=(1/S)(1+{\xi}_{nj})$ for $j \in \mathcal S$, 
$$\mathbb E \left [ \tilde r_n(\bX)- r(\bX)\right]^2 \leq  \frac{2SL^2}{k_n^{\frac{0.75}{S\log 2} (1+\gamma_{n})}} 
+ \left[\sup_{\mathbf x \in [0,1]^d}r^2(\mathbf x)\right] e^{-n/2k_n},$$
where $\gamma_n=\min_{j \in \mathcal S} \xi_{nj}$ tends to $0$ as $n$ tends to infinity.
\end{pro}
This result essentially shows that the rate at which the bias decreases to 0 depends on the number of strong variables, not on $d$. In particular, the quantity ${k_n}^{-(0.75/(S \log 2))(1+\gamma_n)}$ should be compared with the ordinary partitioning estimate bias, which is of the order ${k_n}^{-2/d}$ under the smoothness conditions of Proposition \ref{biais} (see for instance \cite{Laciregressionbook}). In this respect, it is easy to see that ${k_n}^{-(0.75/(S \log 2))(1+\gamma_n)}=\mbox{o}({k_n}^{-2/d})$ as soon as $S\leq \lfloor 0.54 d \rfloor$ ($\lfloor . \rfloor$ is the integer part function). In other words, when the number of active variables is less than (roughly) half of the ambient dimension, the bias of the random forests regression estimate decreases to 0 much faster than the usual rate. The restriction $S\leq \lfloor 0.54 d \rfloor$ is not severe, since in all practical situations we have in mind, $d$ is usually very large with respect to $S$ (this is, for instance, typically the case in modern genome biology problems, where $d$ may be of the order of billion, and in any case much larger than the actual number of active features). Note at last that, contrary to Proposition \ref{variance2}, the term $e^{-n/2k_n}$ prevents the extreme choice $k_n=n$ (about one observation on average in each terminal node). Indeed, an inspection of the proof of Proposition \ref{biais} reveals that this term accounts for the probability that $N_n(\bX,\Theta)$ is precisely 0, i.e., $A_n(\bX,\Theta)$ is empty.
\medskip

\begin{figure}[!h]
\begin{center}
\centering
\includegraphics*[width=12cm,height=9cm]{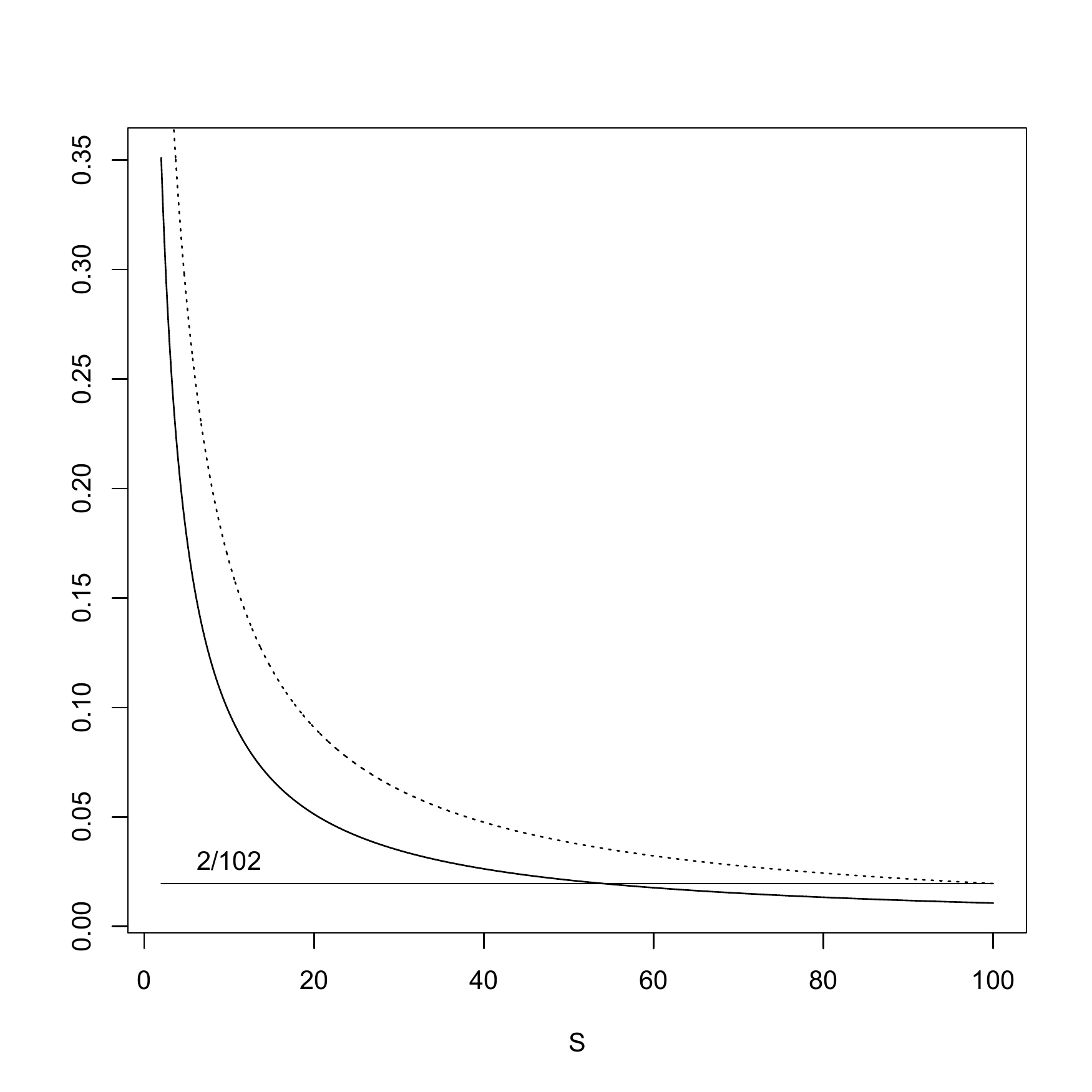}
\caption{\label{fig1} \textsf{{\bf Solid line}: Plot of the function $S\mapsto 0.75/(S\log 2+0.75)$ for $S$ ranging from $2$ to $d=100$.  {\bf Dotted line}: Plot of the minimax rate power $S\mapsto 2/(S+2)$. The horizontal line shows the value of the $d$-dimensional rate power $2/(d+2)\approx 0.0196$.}}
\end{center}
\end{figure}

Recalling the elementary inequality $ze^{-nz} \leq e^{-1}/n$ for $z \in [0,1]$, we may finally join Proposition \ref{variance2} and Proposition \ref{biais} and state our main theorem.
\begin{theo}
\label{gros}
Assume that $\bX$ is uniformly distributed on $[0,1]^d$,  $r^{\star}$ is $L$-Lipschitz on $[0,1]^S$ and, for all $\bx \in \mathbb R^d$,
$$\sigma^2(\bx)=\mathbb V[Y\,|\,\bX=\bx]\leq \sigma^2$$
for some positive constant $\sigma^2$. Then, if $p_{nj}=(1/S)(1+{\xi}_{nj})$ for $j \in \mathcal S$,
letting $\gamma_n=\min_{j \in \mathcal S} \xi_{nj}$, we have
$$
\mathbb E \left [\bar r_n(\bX)-r(\bX) \right]^2  \leq \Xi_n \frac{k_n}{n} +  \frac{2 SL^2}{k_n^{\frac{0.75}{S\log 2} (1+\gamma_{n})}},
$$
where
$$\Xi_n=C\sigma^2\left(\frac{S^2}{S-1}\right)^{S/2d} (1+\xi_n) + 2e^{-1} \left[\sup_{\mathbf x \in [0,1]^d}r^2(\mathbf x)\right]$$
and
$$C=\frac{288}{\pi}\left(\frac{\pi \log 2}{16}\right)^{S/2d}.$$
The sequence $(\xi_n)$ depends on the sequences $\{(\xi_{nj}) : j\in \mathcal S\}$ only and tends to $0$ as $n$ tends to infinity. 
\end{theo}
As we will see in Section 3, it may be safely assumed that the randomization process allows for $\xi_{nj}\log n \to 0$ as $n \to \infty$, for all $j \in \mathcal S$. Thus, under this condition, Theorem \ref{gros} shows that with the optimal choice 
$$k_n \propto n^{1/(1+{\frac{0.75}{S\log 2}})},$$
we get
$$
\mathbb E \left [\bar r_n(\bX)-r(\bX) \right]^2 = \mathcal O \left(n^{ \frac{-0.75}{S\log 2 + 0.75}}\right).$$
This result can be made more precise. Denote by $\mathcal F_S$ the class of $(L,\sigma^2)$-smooth distributions $(\bX,Y)$ such that $\bX$ has uniform distribution on $[0,1]^d$, the regression function $r^{\star}$ is Lipschitz with constant $L$ on $[0,1]^S$ and, for all $\bx \in \mathbb R^d$, $\sigma^2(\bx)=\mathbb V[Y\,|\,\bX=\bx]\leq \sigma^2$. 
\begin{cor}
\label{coro1}
Let
$$\Xi=C\sigma^2\left(\frac{S^2}{S-1}\right)^{S/2d} + 2e^{-1} \left[\sup_{\mathbf x \in [0,1]^d}r^2(\mathbf x)\right]$$
and
$$C=\frac{288}{\pi}\left(\frac{\pi \log 2}{16}\right)^{S/2d}.$$
Then, if $p_{nj}=(1/S)(1+{\xi}_{nj})$ for $j \in \mathcal S$, with $\xi_{nj}\log n\to 0$ as $n \to \infty$, for the choice
$$k_n \propto \left (\frac{L^2}{\Xi}\right)^{1/(1+{\frac{0.75}{S\log 2}})}n^{1/(1+{\frac{0.75}{S\log 2}})},$$
we have
$$\limsup_{nÊ\to \infty} \sup_{(\bX,Y) \in \mathcal F_S}\frac{\mathbb E \left[\bar r_n(\bX)-r(\bX)\right]^2}{\left(\Xi L^{\frac{2S\log 2}{0.75}}\right)^{\frac{0.75}{S\log 2+0.75}}n^{ \frac{-0.75}{S\log 2 + 0.75}}}\leq \Lambda,$$
where $\Lambda$ is a positive constant independent of $r$, $L$ and $\sigma^2$.
\end{cor}
This result reveals the fact that the $L_2$-rate of convergence of $\bar r_n(\bX)$ to $r(\bX)$ depends only on the number $S$ of strong variables, and not on the ambient dimension $d$. The main message of Corollary \ref{coro1} is that if we are able to properly tune the probability sequences $(p_{nj})_{n \geq 1}$ and make them sufficiently fast to track the informative features, then the rate of convergence of the random forests estimate will be of the order $n^{ \frac{-0.75}{S\log 2 + 0.75}}$. This rate is strictly faster than the usual rate $n^{-2/(d+2)}$ as soon as $S\leq \lfloor 0.54d\rfloor$. To understand this point, just recall that the rate $n^{-2/(d+2)}$ is minimax optimal for the class $\mathcal F_d$ (see for example Ibragimov and Khasminskii \cite{IK1, IK2, IK3}), seen as a collection of regression functions over $[0,1]^d$, {\it not} $[0,1]^S$. However, in our setting, the intrinsic dimension of the regression problem is $S$, not $d$, and the random forests estimate cleverly adapts to the sparsity of the problem. As an illustration, Figure \ref{fig1} shows the plot of the function $S\mapsto 0.75/(S\log 2+0.75)$ for $S$ ranging from $2$ to $d=100$.
\medskip

It is noteworthy that the rate of convergence of the $\xi_{nj}$ to 0 (and, consequently, the rate at which the probabilities $p_{nj}$ approach $1/S$ for $j\in \mathcal S$) will eventually depend on the ambient dimension $d$ through the ratio $S/d$. The same is true for the Lipschitz constant $L$ and the factor $\sup_{\bx \in [0,1]^d} r^2(\bx)$ which both appear in Corollary \ref{coro1}. To figure out this remark, remember first that the support of $r$ is contained in $\mathbb R^S$, so that the later supremum (respectively, the Lipschitz constant) is in fact a supremum (respectively, a Lipschitz constant) over $\mathbb R^S$, {\it not} over $\mathbb R^d$. Next, denote by $\mathcal C_p(s)$ the collection of functions $\eta:[0,1]^p\to [0,1]$ for which each derivative of order $s$ satisfies a Lipschitz condition. It is well known that the $\varepsilon$-entropy $\log_2(\mathcal N_{\varepsilon})$ of $\mathcal C_p(s)$ is $\Phi(\varepsilon^{-p/(s+1)})$ as $\varepsilon \downarrow 0$ (Kolmogorov and Tihomirov \cite{KT}), where $a_n=\Phi(b_n)$ means that $a_n=\mathcal O(b_n)$ and $b_n=\mathcal O(a_n)$.  Here we have an interesting interpretation of the dimension reduction phenomenon: Working with Lipschitz functions on $\mathbb R^S$ (that is, $s=0$) is roughly equivalent to working with functions on $\mathbb R^{d}$ for which all $[(d/S)-1]$-th order derivatives are Lipschitz! For example, if $S=1$ and $d=25$, $(d/S)-1=24$ and, as there are $25^{24}$ such partial derivatives in $\mathbb R^{25}$, we note immediately the potential benefit of recovering the ``true'' dimension $S$.
\medskip

{\bf Remark 2}\quad The reduced-dimensional rate $n^{ \frac{-0.75}{S\log 2 + 0.75}}$ is strictly larger than the $S$-dimensional optimal rate $n^{-2/(S+2)}$, which is also shown in Figure \ref{fig1} for $S$ ranging from $2$ to $100$. We do not know whether the latter rate can be achieved by the algorithm. \hfill $\blacksquare$
\medskip

{\bf Remark 3}\quad The optimal parameter $k_n$ of Corollary \ref{coro1} depends on the unknown distribution of $(\bX,Y)$, especially on the smoothness of the regression function and the effective dimension $S$.  To correct this situation, adaptive (i.e.,  data-dependent) choices of $k_n$, such as data-splitting or cross-validation, should preserve the rate of convergence of the estimate.  Another route we may follow is to analyse the effect of bootstrapping the sample before growing the individual trees (i.e., bagging). It is our belief that this procedure should also preserve the rate of convergence, even for overfitted trees ($k_n \approx n$), in the spirit of \cite{BCG}. However, such a study is beyond the scope of the present paper. \hfill $\blacksquare$
\medskip

{\bf Remark 4}\quad For further references, it is interesting to note that Proposition \ref{variance2} (variance term) is a consequence of aggregation, whereas Proposition \ref{biais} (bias term) is a consequence of randomization.
\medskip

It is also stimulating to keep in mind the following analysis, which has been suggested to us by a referee. Suppose, to simplify, that $Y=r(\bX)$ (no-noise regression) and that
$\sum_{i=1}^n W_{ni}(\bX,\Theta)=1$ a.s. In this case, the variance term is 0 and we have
$$
\bar r_n(\bX)=\tilde r_n(\bX)=\sum_{i=1}^n \mathbb E_{\Theta}\left[W_{ni}(\Theta,\bX)\right]Y_i.
$$
Set $\mathbf Z_n=(Y,Y_1, \hdots,Y_n)$.
Then
\begin{align*}
&\mathbb E \left [\bar r_n(\bX)- r(\bX)\right]^2 \nonumber\\
&\quad =\mathbb E \left [ \bar r_n(\bX)- Y\right]^2\nonumber\\
& \quad = \mathbb E \left [ \mathbb E \left [\left ( \bar r_n(\bX)- Y\right)^2 \,|\,\mathbf Z_n  \right]\right]\nonumber\\
&\quad = \mathbb E \left [ \mathbb E \left [ \left (\bar r_n(\bX)- \mathbb E [\bar r_n(\bX)\,|\,\mathbf Z_n]\right)^2\,|\,\mathbf Z_n\right]\right]\nonumber\\
& \qquad + \mathbb E\left [\mathbb E [\bar r_n(\bX)\,|\,\mathbf Z_n]-Y\right]^2\label{leterme}.
\end{align*}
The conditional expectation in the first of the two terms above may be rewritten under the form
\begin{equation*}
\label{a}
\mathbb E \left[\mbox{Cov} \left ( \mathbb E_{\Theta}\left [r_n(\bX,\Theta)\right],\mathbb E_{\Theta'}\left [r_n(\bX,\Theta')\right]\,|\,\mathbf Z_n\right)\right],
\end{equation*}
where $\Theta'$ is distributed as, and independent of, $\Theta$. Attention shows that this last term is indeed equal to
\begin{equation*}
\label{b}
\mathbb E \left[\mathbb E_{\Theta,\Theta'} \mbox{Cov} \left ( r_n(\bX,\Theta),r_n(\bX,\Theta') \,|\, \mathbf Z_n\right)\right]
\end{equation*}
The key observation is that if trees have strong predictive power, then they can be unconditionally strongly correlated while being conditionally weakly correlated. This opens an interesting line of research for the statistical analysis of the bias term, in connection with Amit \cite{Amit2002} and Blanchard \cite{Blanchard} conditional covariance-analysis ideas.
\hfill $\blacksquare$
\section{Discussion}
The results which have been obtained in Section 2 rely on appropriate behavior of the probability sequences $(p_{nj})_{n\geq 1}$, $j=1, \hdots,d$. We recall that these sequences should be in $(0,1)$ and obey the constraints $p_{nj}=(1/S)(1+\xi_{nj})$ for $j\in \mathcal S$ (and $p_{nj}=\xi_{nj}$ otherwise), where the $(\xi_{nj})_{n\geq 1}$ tend to $0$ as $n$ tends to infinity. In other words, at each step of the construction of the individual trees, the random procedure should track and preferentially cut the strong coordinates. In this more informal section, we briefly discuss a random mechanism for inducing such probability sequences. 
\medskip

Suppose, to start with an imaginary scenario, that we already know which coordinates are strong, and which are not. In this ideal case, the random selection procedure described in the introduction may be easily made more precise as follows. A positive integer $M_n$---possibly depending on $n$---is fixed beforehand and the following splitting scheme is iteratively repeated at each node of the tree:
\begin{enumerate}
\item \label{11}Select at random, with replacement, $M_n$ candidate coordinates to split on.
\item \label{22} If the selection is all weak, then choose one at random to split on. If there is more than one strong variable elected, choose one at random and cut.
\end{enumerate}
Within this framework, it is easy to see that each coordinate in
$\mathcal S$ will be cut with the ``ideal'' probability 
$$p_n^{\star}=\frac{1}{S}\left[ 1-\left(1-\frac{S}{d}\right)^{M_n}\right].$$
Though this is an idealized model, it already gives some information about the choice of the parameter $M_n$, which, in accordance with the results of Section 2 (Corollary \ref{coro1}), should satisfy 
$$\left(1-\frac{S}{d}\right)^{M_n}\log n \to 0 \quad \mbox{as } n \to \infty.$$
This is true as soon as
$$M_n\to \infty \quad \mbox{and} \quad \frac{M_n}{\log n} \to \infty \quad \mbox{as } n \to \infty.$$
This result is consistent with the general empirical finding that $M_n$ (called \texttt{mtry} in the R package \texttt{RandomForests}) does not need to be very large (see, for example, Breiman \cite{Bre01}), but not with the widespread belief that $M_n$ should not depend on $n$. Note also that if the $M_n$ features are chosen at random {\it without} replacement, then things are even more simple since, in this case, $p^{\star}_n=1/S$ for all $n$ large enough. 
\medskip

In practice, we have only a vague idea about the size and content of the set $\mathcal S$. However, to circumvent this problem, we may use the observations of an independent second set $\mathcal D'_n$ (say, of the same size as $\mathcal D_n)$ in order to mimic the ideal split probability $p_n^{\star}$. To illustrate this mechanism, suppose---to keep things simple---that the model is linear, i.e.,
$$Y=\sum_{j \in \mathcal S} a_j X^{(j)}+\varepsilon,$$
where $\bX=(X^{(1)}, \hdots, X^{(d)})$ is uniformly distributed over $[0,1]^d$, the $a_j$ are non-zero real numbers, and $\varepsilon$ is a zero-mean random noise, which is assumed to be independent of $\bX$ and with finite variance. Note that, in accordance with our sparsity assumption, $r(\bX)=\sum_{j \in \mathcal S} a_j X^{(j)}$ depends on $\bX_{\mathcal S}$ only. 
\medskip

Assume now that we have done some splitting and arrived at a current set of terminal nodes. Consider any of these nodes, say $A=\prod_{j=1}^d A_j$, fix a coordinate $j\in \{1, \hdots, d\}$, and look at the weighted conditional variance $\mathbb V[Y|X^{(j)} \in A_j]\,\mathbb  P(X^{(j)}\in A_j)$. It is a simple exercise to prove that if $\bX$ is uniform and $j \in \mathcal S$, then the split on the $j$-th side which most decreases the weighted conditional variance is at the midpoint of the node, with a variance decrease equal to $a_j^2/16>0$. On the other hand, if $j\in \mathcal W$,  the decrease of the variance is always 0, whatever the location of the split.
\medskip

On the practical side, the conditional variances are of course unknown, but they may be estimated by replacing the theoretical quantities by their respective sample estimates (as in the CART procedure, see Breiman et al.~\cite[Chapter 8]{Bre01} for a thorough discussion) evaluated on the second sample $\mathcal D'_n$. This suggests the following procedure, at each node of the tree:
\begin{enumerate}
\item \label{one}Select at random, with replacement, $M_n$ candidate coordinates to split on.
\item \label{two}For each of the $M_n$ elected coordinates, calculate the best split, i.e., the split which most decreases the within-node sum of squares on the second sample $\mathcal D'_n$. 
\item \label{three}Select one variable at random among the coordinates which output the best within-node sum of squares decreases, and cut.
\end{enumerate}
This procedure is indeed close to what the random forests algorithm does. The essential difference is that we suppose to have at hand a second sample $\mathcal D'_n$, whereas the original algorithm performs  the search of the optimal cuts on the original observations $\mathcal D_n$. This point is important, since the use of an extra sample preserves the independence of $\Theta$ (the random mechanism) and $\mathcal D_n$ (the training sample). We do not know whether our results are still true if $\Theta$ depends on $\mathcal D_n$ (as in the CART algorithm), but the analysis does not appear to be simple. Note also that, at step \ref{three}, a threshold (or a test procedure, as suggested in Amaratunga et al.~\cite{Amaratunga}) could be used to choose among the most significant variables, whereas the actual algorithm just selects the best one. In fact, depending on the context and the actual cut selection procedure, the informative probabilities $p_{nj}$ ($j\in \mathcal S$) may obey the constraints $p_{nj} \to p_j$ as $n \to \infty$ (thus, $p_j$ is not necessarily equal to $1/S$), where the $p_{j}$ are positive and satisfy $\sum_{j\inÊ\mathcal S}p_{j}=1$. This should not affect the results of the article.
\medskip

This empirical randomization scheme leads to complicate probabilities of cuts which, this time, vary at each node of each tree and are not easily amenable to analysis. Nevertheless, observing that the average number of cases per terminal node is about $n/k_n$, it may be inferred by the law of large numbers that each variable in $\mathcal S$ will be cut with probability
$$p_{nj}\approx \frac{1}{S}\left[ 1-\left(1-\frac{S}{d}\right)^{M_n}\right](1+\zeta_{nj}),$$
where $\zeta_{nj}$ is of the order $\mathcal{O}(k_n/ n)$, a quantity which anyway goes fast to 0 as $n$ tends to infinity. Put differently, for $j \in \mathcal S$,
$$p_{nj}\approx\frac{1}{S}\left(1+\xi_{nj}\right),$$
where $\xi_{nj}$ goes to $0$ and satisfies the constraint $\xi_{nj}\log n \to 0$ as $n$ tends to infinity, provided $k_n\log n /n \to 0$, $M_n \to \infty$ and $M_n/\log n \to \infty$. This is coherent with the requirements of Corollary \ref{coro1}. We realize however that this is a rough approach, and that more theoretical work is needed here to fully understand the mechanisms involved in CART and Breiman's original randomization process.
\medskip

It is also noteworthy that random forests use the so-called out-of-bag samples (i.e., the bootstrapped data which are not used to fit the trees) to construct a variable importance criterion, which measures the prediction strength of each feature (see, e.g., Genuer et al.~\cite{Genuer2}). As far as we are aware, there is to date no systematic mathematical study of this criterion. It is our belief that such a study would greatly benefit from the sparsity point of view developed in the present paper, but is unfortunately much beyond its scope. Lastly, it would also be interesting to work out and extend our results to the context of unsupervised learning of trees. A good route to follow with this respect is given by the strategies outlined in Amit and Geman \cite[Section 5.5]{Amit}.
\section{A small simulation study}
\begin{figure}[h]
\begin{center}
\centering
\includegraphics*[width=8cm,height=8cm]{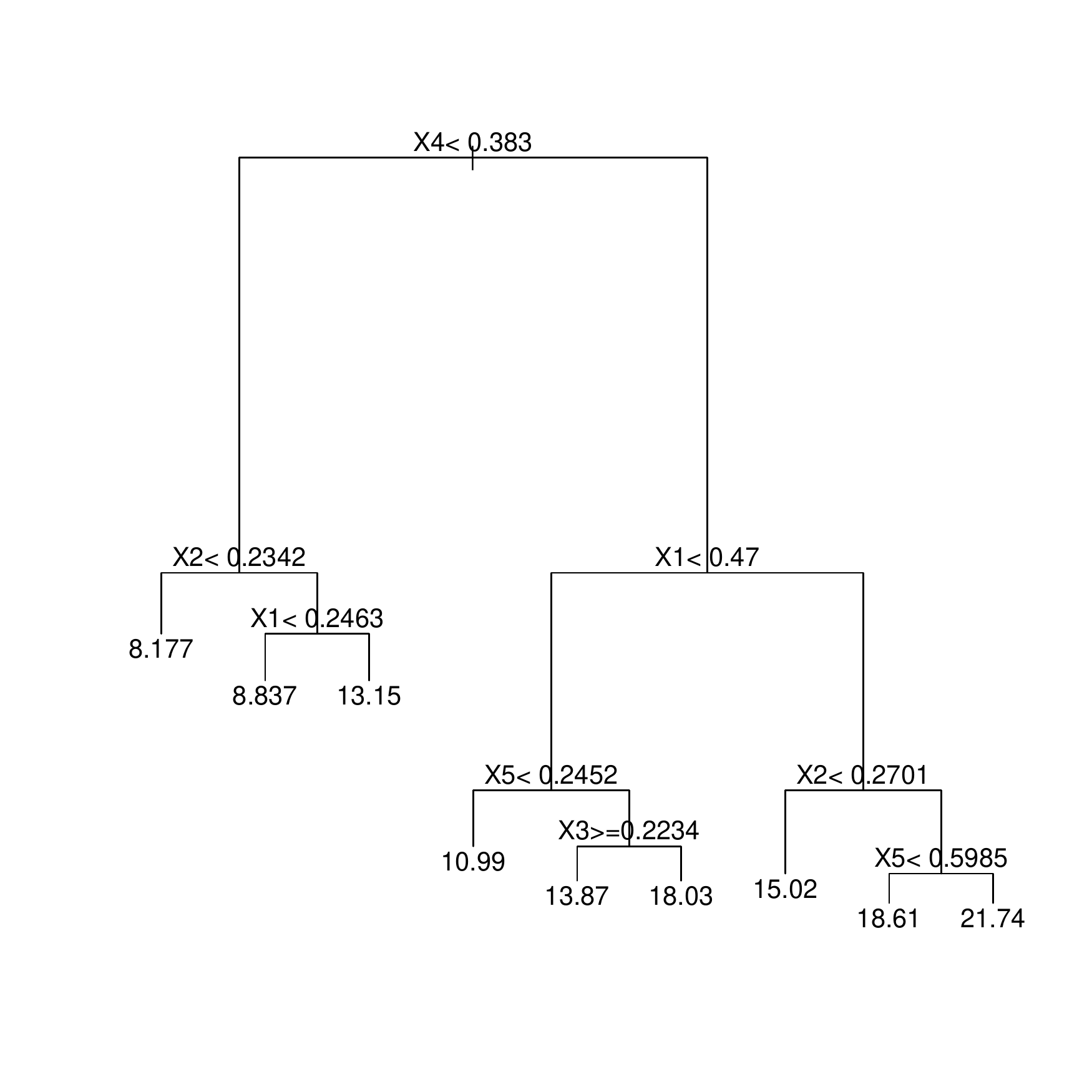}
\caption{\label{Tree}\textsf{The tree used as regression function in the model {\bf Tree}.}}
\end{center}
\end{figure}

Even though the first vocation of the present paper is theoretical, we offer in this short section some experimental results on synthetic data. Our aim is not to provide a thorough practical study of the random forests method, but rather to illustrate the main ideas of the article. As for now, we let $\mathcal{U}([0,1]^d)$ (respectively, $\mathcal N(0,1)$) be the uniform distribution over $[0,1]^d$ (respectively, the standard Gaussian distribution). Specifically, three models were tested:
\begin{itemize}
\item[1.][{\bf Sinus}] For $\bx \in [0,1]^d$, the regression function takes the form
$$r(\bx)= 10 \sin(10 \pi x^{(1)}).$$ 
We let $Y = r(\bX) + \varepsilon$ and $\bX\sim\mathcal{U}([0,1]^d)$ ($d\geq 1$), with $\varepsilon\sim\mathcal{N}(0,1)$.
 \item[2.][{\bf Friedman \#1}] This is a model proposed in Friedman \cite{Fri91}. Here, 
 $$r(\bx)=10 \sin(\pi x^{(1)} x^{(2)}) + 20(x^{(3)}-.05)^2 + 10 x^{(4)} + 5 x^{(5)}$$
 and $Y = r(\bX) + \varepsilon$, where $\bX\sim\mathcal{U}([0,1]^d)$ ($d\geq 5$) and $\varepsilon\sim\mathcal{N}(0,1)$.
 \item[3.][{\bf Tree}] In this example, we let $Y = r(\bX) + \varepsilon$, where $\bX\sim\mathcal{U}([0,1]^d)$ ($d\geq 5$), $\varepsilon\sim\mathcal{N}(0,1)$ and the function $r$ has itself a tree structure. This tree-type function, which is shown in Figure \ref{Tree}, involves only five variables.  
 \end{itemize}
We note that, although the ambient dimension $d$ may be large, the effective dimension of model 1 is $S=1$, whereas model 2 and model 3 have $S=5$. In other words, $\mathcal S=\{1\}$ for model 1, whereas $\mathcal S=\{1, \hdots, 5\}$ for model 2 and model 3. Observe also that, in our context, the model {\bf Tree} should be considered as a ``no-bias'' model, on which the random forests algorithm is expected to perform well.
\medskip

\begin{figure}[h]
\begin{center}
\centering
\includegraphics[width=12cm,height=12cm]{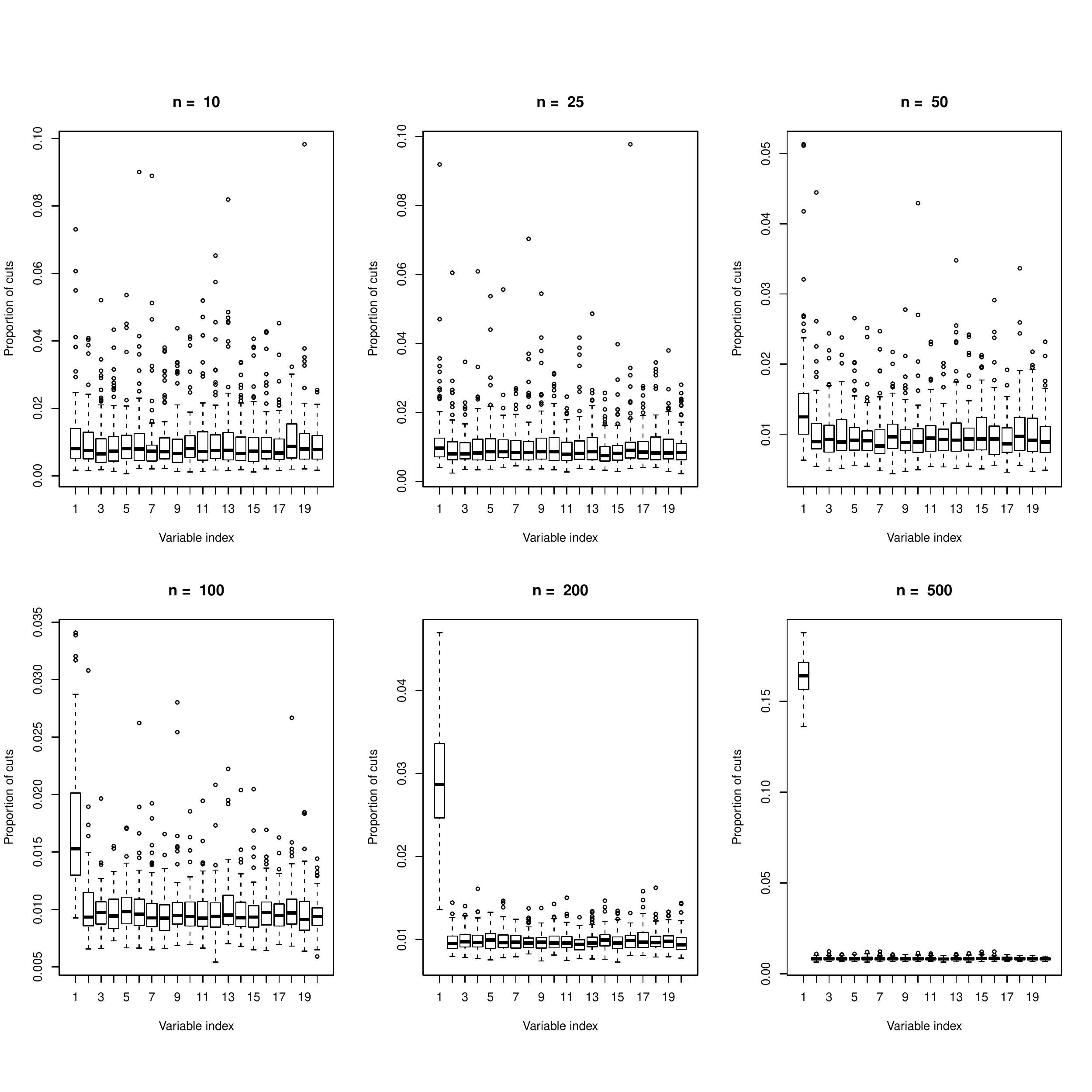}
\caption{\label{F1}\textsf{Boxplots of the empirical probabilities of cuts for model {\bf Sinus} ($\mathcal S=\{1\}$).}}
\end{center}
\end{figure}

\begin{figure}[h]
\begin{center}
\centering
\includegraphics[width=12cm,height=12cm]{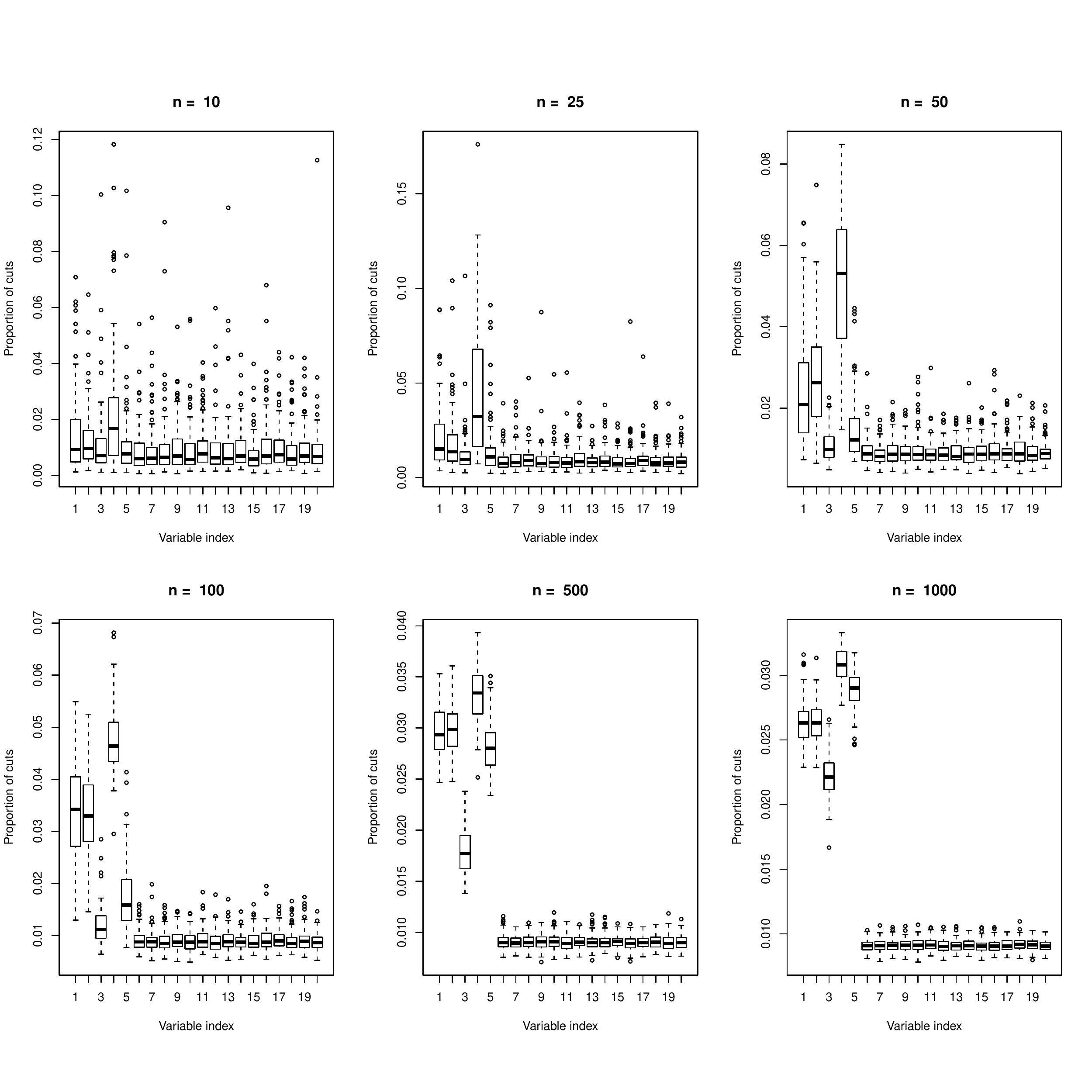}
\caption{\label{F2}\textsf{Boxplots of the empirical probabilities of cuts for model {\bf Friedman \#1} ($\mathcal S=\{1, \hdots, 5\}$).}}
\end{center}
\end{figure}

\begin{figure}[h]
\begin{center}
\centering
\includegraphics[width=12cm,height=12cm]{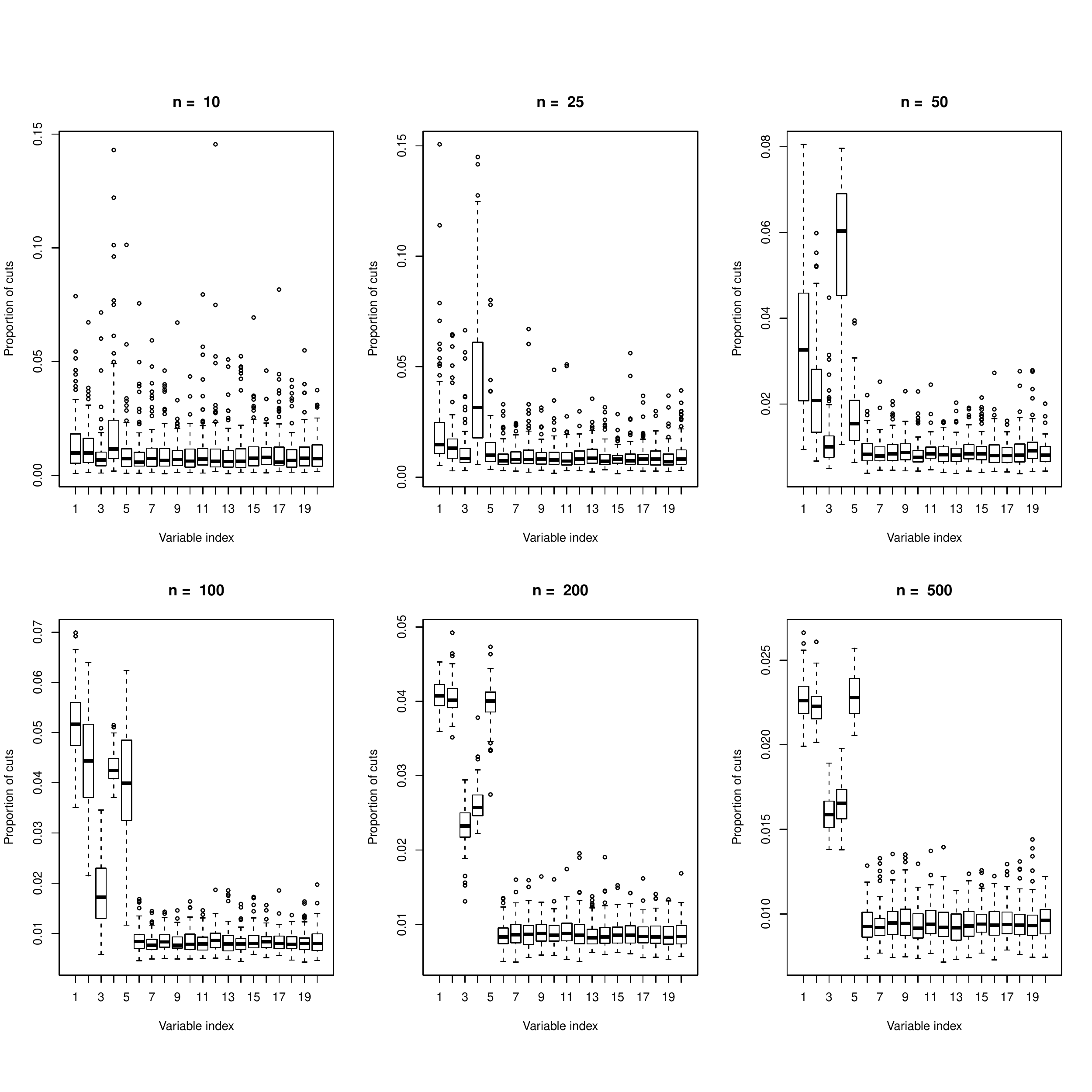}
\caption{\label{F3}\textsf{Boxplots of the empirical probabilities of cuts for model {\bf Tree} ($\mathcal S=\{1, \hdots, 5\}$).}}
\end{center}
\end{figure}

In a first series of experiments, we let $d=100$ and, for each of the three models and different values of the sample size $n$, we generated a learning set of size $n$ and fitted a forest ($10\,000$ trees) with $\texttt{mtry} = d$. For $j=1, \hdots, d$, the ratio (number of times the $j$-th coordinate is split)/(total number of splits over the forest) was evaluated, and the whole experiment was repeated 100 times. Figure \ref{F1}, Figure \ref{F2} and Figure \ref{F3} report the resulting boxplots for each of the first twenty variables and different values of $n$. These figures clearly enlighten the fact that, as $n$ grows, the probability of cuts does concentrate on the informative variables only and support the assumption that $\xi_{nj}\to0$ as $n\to \infty$ for each $j\in \mathcal S$. 
\medskip

Next, in a second series of experiments, for each model, for different values of $d$ and for sample sizes $n$ ranging from 10 to 1000, we generated a learning set of size $n$, a test set of size $50\,000$ and evaluated the mean squared error (MSE) of the random forests (RF) method  via the Monte Carlo approximation
\begin{equation*}
\mbox{MSE} \approx \frac {1}{50\,000} \sum_{j=1}^{50\,000} \left[\mbox{RF}(\mbox{test data \#j}) - r(\mbox{test data \#j}) \right]^2.
\end{equation*}
All results were averaged over 100 data sets. The random forests algorithm was performed with the parameter \texttt{mtry} automatically tuned by the R package \texttt{RandomForests}, $1000$ random trees and the minimum node size set to $5$ (which is the default value for regression). Besides, in order to compare the ``true'' algorithm with the approximate model discussed in the present document, an alternative method was also tested. This auxiliary algorithm has characteristics which are identical to the original ones (same \texttt{mtry}, same number of random trees), {\it with the notable difference that now the maximum number of nodes is fixed beforehand}. For the sake of coherence, since the minimum node size is set to 5 in the \texttt{RandomForests} package, the number of terminal nodes in the custom algorithm was calibrated to $\lceil n/5\rceil$. It must be stressed that the essential difference between the standard random forests algorithm and the alternative one is that the number of cases in the final leaves is fixed in the former, whereas the latter assumes a fixed number of terminal nodes. In particular, in both algorithms, cuts are performed using the actual sample, just as CART does. To keep things simple, no data-splitting procedure has been incorporated in the modified version.
\medskip

\begin{figure}[h]
\begin{center}
\centering
\includegraphics[width=12cm,height=9cm]{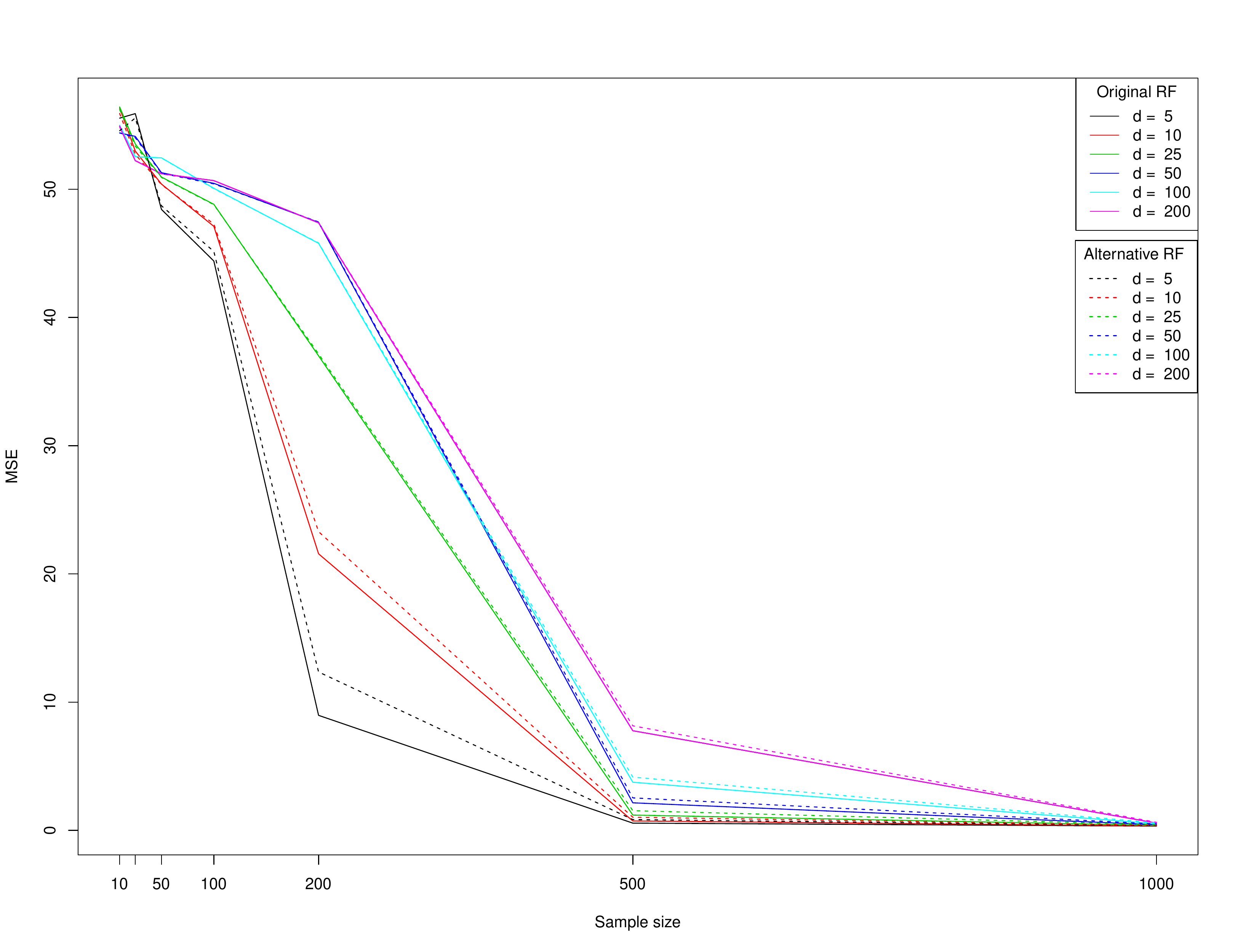}
\caption{\label{F4}\textsf{Evolution of the MSE for model {\bf Sinus} ($S=1$).}}
\end{center}
\end{figure}

\begin{figure}[h]
\begin{center}
\centering
\includegraphics[width=12cm,height=9cm]{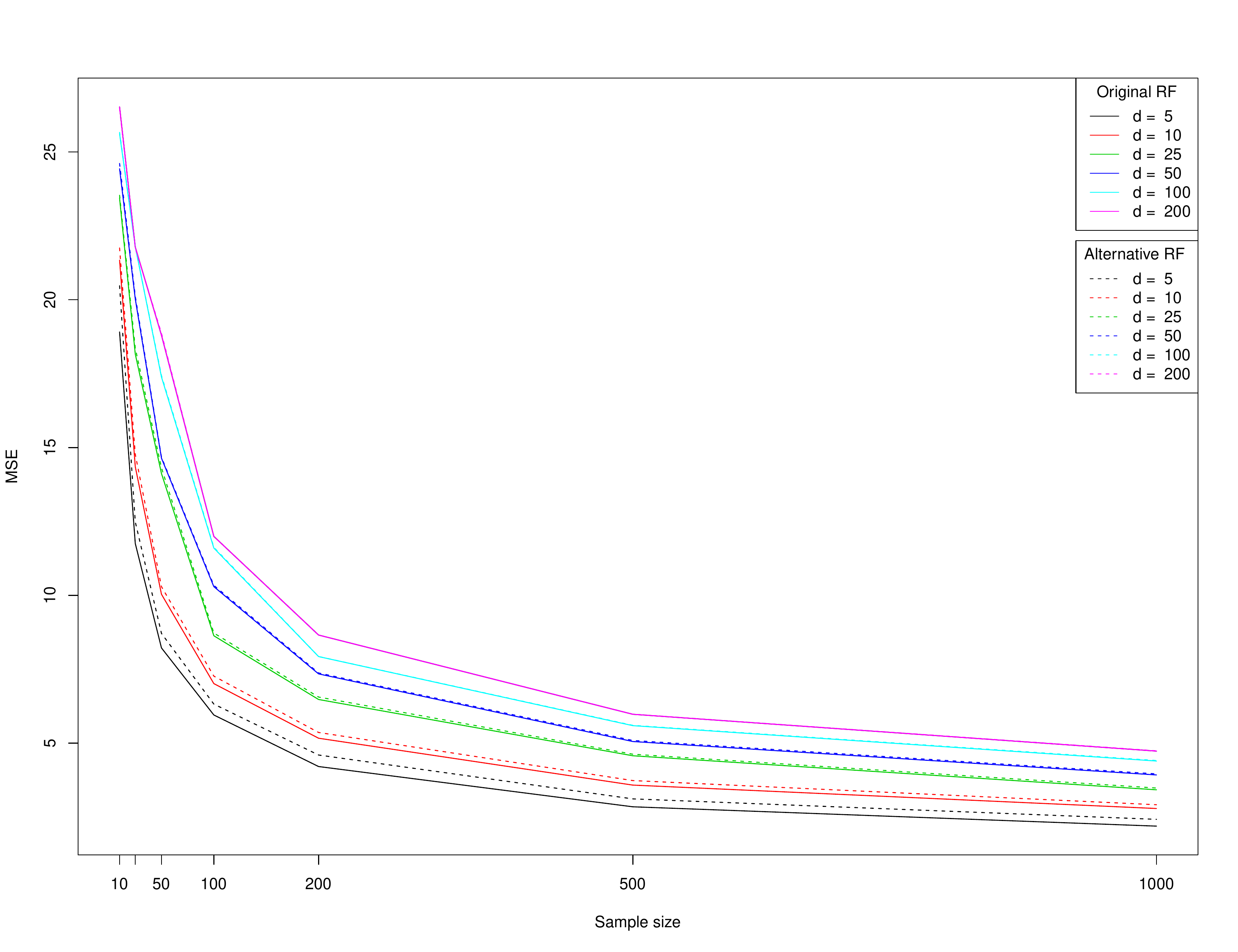}
\caption{\label{F5}\textsf{Evolution of the MSE for model {\bf Friedman \#1} ($S=5$).}}
\end{center}
\end{figure}

\begin{figure}[h]
\begin{center}
\centering
\includegraphics[width=12cm,height=9cm]{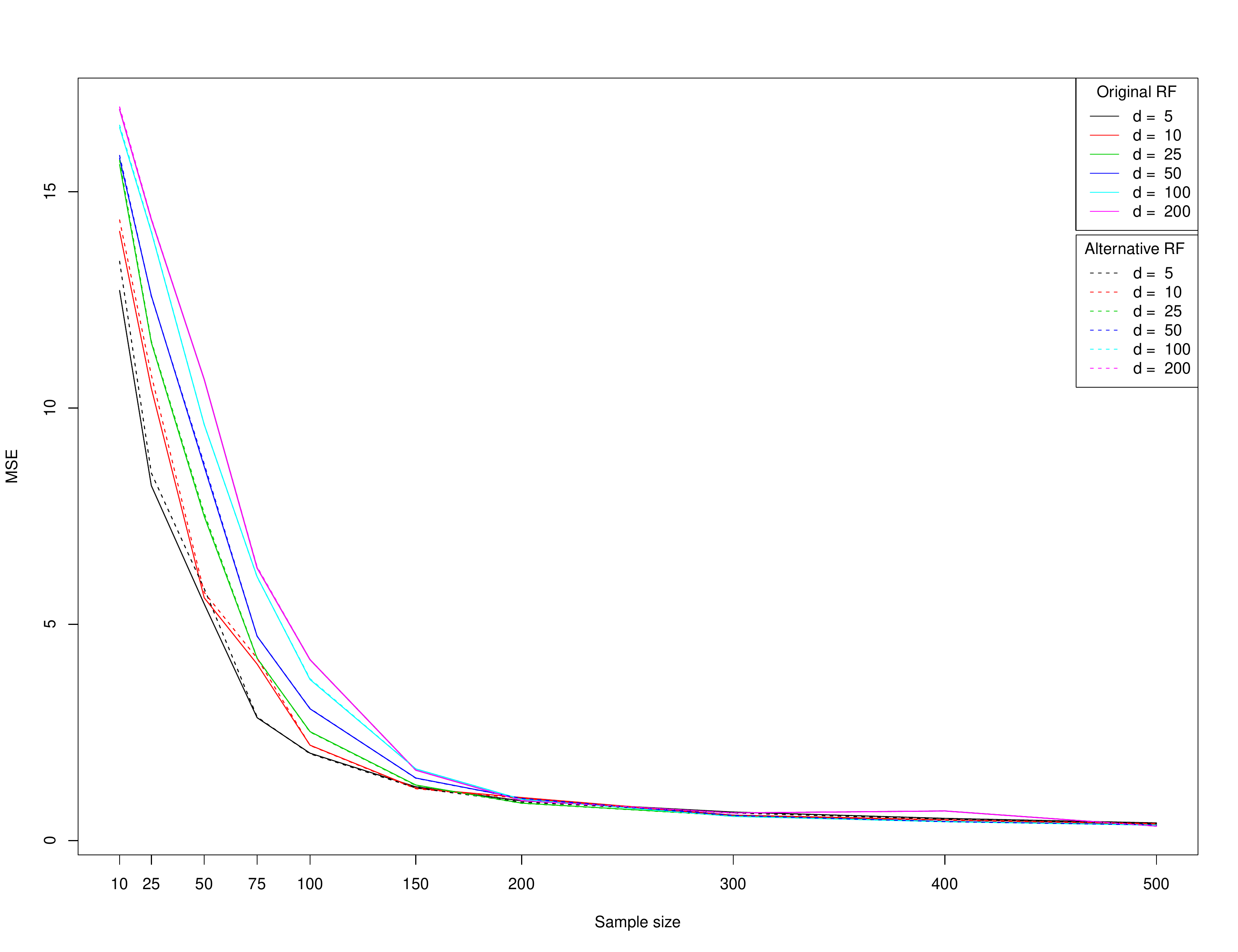}
\caption{\label{F6}\textsf{Evolution of the MSE for model {\bf Tree} ($S=5$).}}
\end{center}
\end{figure}

Figure \ref{F4}, Figure \ref{F5} and Figure \ref{F6} illustrate the evolution of the MSE value with respect to $n$ and $d$, for each model and the two tested procedures. First, we note that the overall performance of the alternative method is very similar to the one of the original algorithm. This confirms our idea that the model discussed in the present paper is a good approximation of the authentic Breiman's forests. Next, we see that for a sufficiently large $n$, the capabilities of the forests are nearly independent of $d$, in accordance with the idea that the (asymptotic) rate of convergence of the method should only depend on the ``true'' dimensionality $S$ (Theorem \ref{gros}). Finally, as expected, it is noteworthy that both algorithms perform well on the third model, which has been precisely designed for a tree-structured predictor.
\section{Proofs}
Throughout this section, we will make repeated use of the following two facts. 
 \begin{fact}
 \label{FAIT1}
Let $K_{nj}(\bX,\Theta)$ be the number of times the terminal node $A_n(\bX,\Theta)$ is split on the $j$-th coordinate ($j=1, \hdots, d$). Then, conditionally on $\bX$, $K_{nj}(\bX,\Theta)$ has binomial distribution with parameters $\lk$ and $p_{nj}$ (by independence of $\bX$ and $\Theta$). Moreover, by construction,
$$\sum_{j=1}^d K_{nj}(\bX,\Theta)=\lk.$$
\end{fact}
Recall that we denote by $N_n(\bX,\Theta)$ the number of data points falling in the same cell as $\bX$, i.e.,
$$
N_n(\bX,\Theta)= \sum_{i=1}^n \mathbf 1_{[\bX_i\in A_n(\bX,\Theta)]}.$$
Let $\lambda$ be the Lebesgue measure on $[0,1]^d$.
\begin{fact}
 \label{FAIT2}
By construction, 
$$\lambda \left (A_n(\bX,\Theta)\right)=2^{-\lk}.$$
In particular, if $\bX$ is uniformly distributed on $[0,1]^d$, then the distribution of $N_n(\bX,\Theta)$ conditionally on $\bX$ and $\Theta$ is binomial with parameters $n$ and $2^{-\lk}$ (by independence of the random variables $\bX, \bX_1, \hdots, \bX_n, \Theta$).
\end{fact}

{\bf Remark 5}\quad If $\bX$ is not uniformly distributed but has a probability density $f$ on $[0,1]^d$, then, conditionally on $\bX$ and $\Theta$,  $N_n(\bX,\Theta)$ is binomial with parameters $n$ and $\mathbb P(\bX_1 \in A_n(\bX,\Theta)\,|\,\bX,\Theta)$. If $f$ is bounded from above and from below, this probability is of the order $\lambda \left (A_n(\bX,\Theta)\right)=2^{-\lk}$, and the whole approach can be carried out without difficulty. On the other hand, for more general densities, the binomial probability depends on $\bX$, and this makes the analysis significantly harder. \hfill $\blacksquare$
\subsection{Proof of Theorem \ref{convergence}} Observe first that, by Jensen's inequality,
\begin{align*}
\mathbb E \left [\bar r_n(\bX)-r(\bX)\right]^2& =\mathbb E \left [ \mathbb E_{\Theta} \left [r_n(\bX,\Theta)-r(\bX)\right]\right]^2\\
& \leq \mathbb E \left [ r_n(\bX,\Theta)-r(\bX)\right]^2.
\end{align*}
A slight adaptation of Theorem 4.2 in Gy\"orfi et al.~\cite{Laciregressionbook} shows that $\bar r_n$ is consistent if both
$\mbox{diam}(A_n(\bX,\Theta)) \to 0$ in probability and 
$N_n(\bX,\Theta) \to \infty$ in probability. 
\medskip

Let us first prove that $N_n(\bX,\Theta) \to \infty$ in probability. To see this, consider the random tree partition defined by $\Theta$, which has by construction exactly $2^{\lk}$ rectangular cells, say $A_1,\ldots,A_{2^{\lk}}$. Let $N_1,\ldots,N_{2^{\lk}}$ denote the number of observations among $\bX,\bX_1,\ldots,\bX_n$ falling in these $2^{\lk}$ cells, and let $\mathcal C=\{\bX,\bX_1,\ldots,\bX_n\}$ denote the set of positions of these $n+1$ points. Since these points are independent and identically distributed, fixing the set $\mathcal C$ and $\Theta$, the conditional probability that $\bX$ falls in the $\ell$-th cell equals $N_{\ell}/(n+1)$. Thus, for every fixed $M\geq 0$,
\begin{align*}
  \mathbb P \left(N_n(\bX,\Theta) < M\right) & = \mathbb E \left[\mathbb P \left(N_n(\bX,\Theta) < M\,|\, \mathcal C,\Theta\right) \right]  \\
&= \mathbb E \left[ \sum_{\ell=1, \hdots, 2^{\lk} : N_{\ell} < M} \frac{N_{\ell}}{n+1} \right]\\
& \leq \frac{M2^{\lk}}{n+1}\\
& \leq \frac{2Mk_n}{n+1},
\end{align*}
which converges to 0 by our assumption on $k_n$.
\medskip

It remains to show that $\mbox{diam}(A_n(\bX,\Theta)) \to 0$ in probability. To this aim, let $V_{nj}(\bX,\Theta)$ be the size of the $j$-th dimension of the rectangle
containing $\bX$. Clearly, it suffices to show that $V_{nj} (\bX,\Theta) \to 0$ in probability for all $j=1, \hdots, d$. To this end, note that 
$$V_{nj} (\bX,\Theta)\stackrel{\mathcal D}{=}2^{-K_{nj}(\bX,\Theta)},$$
where, conditionally on $\bX$, $K_{nj}(\bX,\Theta)$ has a binomial $\mathcal B(\lk,p_{nj})$ distribution, representing the number of
times the box containing $\bX$ is split along the $j$-th coordinate (Fact \ref{FAIT1}). Thus
\begin{align*}
\mathbb E \left [V_{nj} (\bX,\Theta) \right]& =\mathbb E \left[2^{-K_{nj}(\bX,\Theta)}\right]\\
& =\mathbb E \left[\mathbb E \left[2^{-K_{nj}(\bX,\Theta)}\,|\,\bX\right]\right]\\
& =(1-p_{nj}/2)^{\lk},
\end{align*}
which tends to $0$ as $p_{nj}\log k_n \to \infty$.
\subsection{Proof of Proposition \ref{variance2}} Recall that
$$\bar r_n(\bX)=\sum_{i=1}^n \mathbb E_{\Theta}\left [W_{ni}(\bX,\Theta)\right] Y_i,$$
where
$$W_{ni}(\bX,\Theta)=\frac{\mathbf 1_{[\bX_i \in A_n(\bX,\Theta)]}}{N_n(\bX,\Theta)}\,\mathbf 1_{\mathcal E_n(\bX,\Theta)}$$
and
$$\mathcal E_n=\left [N_n(\bX,\Theta) \neq 0 \right ].$$
Similarly,
$$\tilde r_n(\bX)= \sum_{i=1}^n\mathbb E_{\Theta}\left [W_{ni}(\bX,\Theta)\right] r({\bX}_i).$$
We have
\begin{align*}
\mathbb E \left [\bar r_n(\bX)-\tilde r_n(\bX) \right]^2 & =  \mathbb E \left [\sum_{i=1}^n \mathbb E_{\Theta}\left [W_{ni}(\bX,\Theta)\right]\left(Y_i-r(\bX_i)\right)\right]^2\\
& =  \mathbb E \left [\sum_{i=1}^n \mathbb E_{\Theta}^2\left [W_{ni}(\bX,\Theta)\right]\left(Y_i-r(\bX_i)\right)^2\right]\\
& \quad (\mbox{the cross terms are } 0 \mbox{ since } \mathbb E [Y_i|\bX_i]=r(\bX_i))\\
&  =\mathbb E \left [\sum_{i=1}^n \mathbb E_{\Theta}^2\left [W_{ni}(\bX,\Theta)\right]\sigma^2(\bX_i)\right]\\
&\leq \sigma^2 \mathbb E \left[  \sum_{i=1}^n \mathbb E^2_{\Theta} \left[ W_{ni}(\bX,\Theta) \right]  \right]\\
&= n \sigma^2 \mathbb E \left[\mathbb E^2_{\Theta} \left[ W_{n1}(\bX,\Theta) \right]\right],
\end{align*}
where we used a symmetry argument in the last equality. Observe now that
\begin{align*}
\mathbb E^2_{\Theta} \left[ W_{n1}(\bX,\Theta) \right] &= \mathbb E_{\Theta} \left[ W_{n1}(\bX,\Theta) \right] \mathbb E_{\Theta'} \left[ W_{n1}(\bX,\Theta') \right]\\
& \qquad (\mbox{where } \Theta' \mbox{ is distributed as, and independent of, } \Theta)\\
& \quad =\mathbb E_{\Theta,\Theta'} \left [ W_{n1}(\bX,\Theta)W_{n1}(\bX,\Theta')\right]\\
& \quad =\mathbb E_{\Theta,\Theta'} \left[\frac{ \mathbf 1_{[\bX_1 \in A_n(\bX,\Theta)]} \mathbf 1_{[\bX_1 \in A_n(\bX,\Theta')]}}{N_n(\bX,\Theta)N_n(\bX,\Theta')}\,\mathbf 1_{\mathcal E_n(\bX,\Theta)}\mathbf 1_{\mathcal E_n(\bX,\Theta')}\right]\\
& \quad = \mathbb E_{\Theta,\Theta'} \left[\frac{ \mathbf 1_{[\bX_1 \in A_n(\bX,\Theta)\cap  A_n(\bX,\Theta')]}}{N_n(\bX,\Theta)N_n(\bX,\Theta')}\,\mathbf 1_{\mathcal E_n(\bX,\Theta)}\mathbf 1_{\mathcal E_n(\bX,\Theta')}\right].
\end{align*}
Consequently,
$$\mathbb E \left [\bar r_n(\bX)-\tilde r_n(\bX) \right]^2 \leq n \sigma^2\mathbb E \left[\frac{ \mathbf 1_{[\bX_1 \in A_n(\bX,\Theta)\cap  A_n(\bX,\Theta')]}}{N_n(\bX,\Theta)N_n(\bX,\Theta')}\,\mathbf 1_{\mathcal E_n(\bX,\Theta)}\mathbf 1_{\mathcal E_n(\bX,\Theta')}\right].$$
Therefore
\begin{align*}
& \mathbb E \left [\bar r_n(\bX)-\tilde r_n(\bX) \right]^2 \\
& \quad \leq n \sigma^2 \mathbb E \left[\frac{ \mathbf 1_{[\bX_1 \in A_n(\bX,\Theta)\cap  A_n(\bX,\Theta')]}}{\left(1+\sum_{i =2}^n \mathbf 1_{[\bX_i \in A_n(\bX,\Theta)]}\right)\left(1+\sum_{i=2}^n \mathbf 1_{[\bX_i \in A_n(\bX,\Theta')]}\right)}\right]\\
& \quad = n \sigma^2 \mathbb E \Bigg[\mathbb E \Bigg [\frac{ \mathbf 1_{[\bX_1 \in A_n(\bX,\Theta)\cap  A_n(\bX,\Theta')]}}{\left(1+\sum_{i =2}^n \mathbf 1_{[\bX_i \in A_n(\bX,\Theta)]}\right)}\\
& \quad \qquad \qquad \qquad \times \frac{1}{\left(1+\sum_{i =2}^n \mathbf 1_{[\bX_i \in A_n(\bX,\Theta')]}\right)}\,|\, \bX,\bX_1,\Theta,\Theta' \Bigg]\Bigg]\\
& \quad = n \sigma^2 \mathbb E \Bigg[ \mathbf 1_{[\bX_1 \in A_n(\bX,\Theta)\cap  A_n(\bX,\Theta')]}\mathbb E \Bigg[\frac{ 1}{\left(1+\sum_{i =2}^n \mathbf 1_{[\bX_i \in A_n(\bX,\Theta)]}\right)}\\
& \quad \qquad \qquad \qquad \times \frac{1}{\left(1+\sum_{i =2}^n \mathbf 1_{[\bX_i \in A_n(\bX,\Theta')]}\right)}\,|\, \bX,\bX_1,\Theta,\Theta' \Bigg]\Bigg]\\
& \quad = n \sigma^2 \mathbb E \Bigg[ \mathbf 1_{[\bX_1 \in A_n(\bX,\Theta)\cap  A_n(\bX,\Theta')]}\mathbb E \Bigg[\frac{ 1}{\left(1+\sum_{i =2}^n \mathbf 1_{[\bX_i \in A_n(\bX,\Theta)]}\right)}\\
& \quad \qquad \qquad \qquad \times \frac{1}{\left(1+\sum_{i =2}^n \mathbf 1_{[\bX_i \in A_n(\bX,\Theta')]}\right)}\,|\, \bX,\Theta,\Theta' \Bigg]\Bigg]
\end{align*}
by the independence of the random variables $\bX, \bX_1, \hdots, \bX_n, \Theta, \Theta'$. Using the Cauchy-Schwarz inequality, the above conditional expectation can be upper bounded by
\begin{align*}
& {\mathbb E^{1/2} \left [ \frac{1}{\left(1+\sum_{i =2}^n \mathbf 1_{[\bX_i \in A_n(\bX,\Theta)]}\right)^2}\,|\, \bX,\Theta\right ]} \\
& \qquad  \times {\mathbb E^{1/2} \left [ \frac{1}{\left(1+\sum_{i =2}^n \mathbf 1_{[\bX_i \in A_n(\bX,\Theta')]}\right)^2}\,|\, \bX,\Theta'\right ]}\\
& \leq \frac{3\times2^{2\lk}}{n^2}\\
& \quad (\mbox{by Fact \ref{FAIT2} and technical Lemma \ref{bosq}})\\
& \leq \frac{12k_n^2}{n^2}.
\end{align*}
It follows that
\begin{align} 
\mathbb E \left [\bar r_n(\bX)-\tilde r_n(\bX) \right]^2 &\leq  \frac{12\sigma^2k_n^2}{n}\,\mathbb E \left[ \mathbf 1_{[\bX_1 \in A_n(\bX,\Theta)\cap  A_n(\bX,\Theta')]}\right]\nonumber\\
& =\frac{12\sigma^2k_n^2}{n}\,\mathbb E \left [\mathbb E_{\bX_1} \left[ \mathbf 1_{[\bX_1 \in A_n(\bX,\Theta)\cap  A_n(\bX,\Theta')]}\right]\right]\nonumber\\
& =\frac{12\sigma^2k_n^2}{n}\,\mathbb E \left [\mathbb P_{\bX_1} \left(\bX_1 \in A_n(\bX,\Theta)\cap  A_n(\bX,\Theta')\right)\right]\label{bacG}.
\end{align}
Next, using the fact that $\bX_1$ is uniformly distributed over $[0,1]^d$, we may write
\begin{align*}
\mathbb P_{\bX_1} \left(\bX_1 \in A_n(\bX,\Theta)\cap  A_n(\bX,\Theta')\right) &= \lambda \left(A_n(\bX,\Theta)\cap  A_n(\bX,\Theta')\right)\\
&= \prod_{j=1}^d \lambda \left(A_{nj}(\bX,\Theta)\cap  A_{nj}(\bX,\Theta')\right),
\end{align*}
where 
$$A_n(\bX,\Theta)=\prod_{j=1}^d A_{nj}(\bX,\Theta)\quad \mbox{and}\quad A_n(\bX,\Theta')=\prod_{j=1}^d A_{nj}(\bX,\Theta').$$
On the other hand, we know (Fact \ref{FAIT1}) that, for all $j=1, \hdots, d$,
$$\lambda \left(A_{nj}(\bX,\Theta)\right)\stackrel{\mathcal D}{=}2^{-K_{nj}(\bX,\Theta)},$$
where, conditionally on $\bX$, $K_{nj}(\bX,\Theta)$ has a binomial $\mathcal B(\lk,p_{nj})$ distribution and, similarly,
$$\lambda \left(A_{nj}(\bX,\Theta')\right)\stackrel{\mathcal D}{=}2^{-K'_{nj}(\bX,\Theta')},$$
where, conditionally on $\bX$, $K'_{nj}(\bX,\Theta')$ is binomial $\mathcal B(\lk,p_{nj})$ and independent of $K_{nj}(\bX,\Theta)$. In the rest of the proof, to lighten notation, we write $K_{nj}$ and $K'_{nj}$ instead of $K_{nj}(\bX,\Theta)$ and $K'_{nj}(\bX,\Theta')$, respectively. Clearly,
\begin{align*}
\lambda \left(A_{nj}(\bX,\Theta)\cap  A_{nj}(\bX,\Theta')\right) & {\leq} 2^{-\max(K_{nj},K'_{nj})}\\
& =2^{-K'_{nj}}2^{-(K_{nj}-K'_{nj})_+}
\end{align*}
and, consequently,
$$ \prod_{j=1}^d \lambda \left(A_{nj}(\bX,\Theta)\cap  A_{nj}(\bX,\Theta')\right) {\leq}2^{-\lk} \prod_{j=1}^d 2^{-(K_{nj}-K'_{nj})_+}$$
(since, by Fact \ref{FAIT1}, $\sum_{j=1}^d K_{nj}=\lk$). Plugging this inequality into (\ref{bacG}) and applying H\"older's inequality, we obtain
\begin{align*}
\mathbb E \left [\bar r_n(\bX)-\tilde r_n(\bX) \right]^2 &\leq  \frac{12\sigma^2k_n}{n} \mathbb E\left [ \prod_{j=1}^d 2^{-(K_{nj}-K'_{nj})_+}\right]\\
& =  \frac{12\sigma^2k_n}{n} \mathbb E\left[\mathbb E\left [ \prod_{j=1}^d 2^{-(K_{nj}-K'_{nj})_+}\,|\,\bX\right]\right]\\
& \leq  \frac{12\sigma^2k_n}{n} \mathbb E \left[\prod_{j=1}^d \mathbb E^{1/d} \left [2^{-d(K_{nj}-K'_{nj})_+}\,|\,\bX\right]\right].
\end{align*}
Each term in the product may be bounded by technical Proposition \ref{corocauchy}, and this leads to
\begin{align*}
\mathbb E \left [\bar r_n(\bX)-\tilde r_n(\bX) \right]^2 &  \leq \frac{288\sigma^2k_n}{\pi n} \prod_{j=1}^d \min \left (1, \left [\frac{\pi}{16\lk p_{nj}(1-p_{nj})}\right]^{1/2d}\right)\\
& \leq \frac{288\sigma^2k_n}{\pi n} \prod_{j=1}^d \min \left (1, \left [\frac{\pi\log 2}{16(\log k_n)p_{nj}(1-p_{nj})}\right]^{1/2d}\right).
\end{align*}
Using the assumption on the form of the $p_{nj}$, we finally conclude that
$$\mathbb E \left [\bar r_n(\bX)-\tilde r_n(\bX) \right]^2  \leq C\sigma^2\left(\frac{S^2}{S-1}\right)^{S/2d}\left(1+\xi_n\right)\frac{k_n}{n(\log k_n)^{S/2d}},$$
where
$$C=\frac{288}{\pi}\left(\frac{\pi \log 2}{16}\right)^{S/2d}$$
and
$$1+\xi_n=\prod_{j \in \mathcal S} \left [\left( 1+ \xi_{nj} \right)^{-1} \left (1- \frac{\xi_{nj}}{S-1} \right)^{-1}\right]^{1/2d}.$$
Clearly, the sequence $(\xi_n)$, which depends on the $\{(\xi_{nj}) : j\in \mathcal S\}$ only, tends to $0$ as $n$ tends to infinity.
\subsection{Proof of Proposition \ref{biais}}We start with the decomposition
\begin{align*}
&\mathbb E \left [ \tilde r_n(\bX)- r(\bX)\right]^2 \\
& \quad = \mathbb E \Bigg [  \sum_{i=1}^n  \mathbb E_{\Theta} \left [W_{ni}(\bX,\Theta)\right] \left( r(\bX_i)-r(\bX)\right)\\
& \quad \qquad \qquad +\left(\sum_{i=1}^n  \mathbb E_{\Theta} \left [W_{ni}(\bX,\Theta)\right]-1\right) r(\bX)\Bigg]^2\\
& \quad \mbox = \mathbb E\left[ \mathbb E_{\Theta} \left [ \sum_{i=1}^n W_{ni}(\bX,\Theta)\left( r(\bX_i)-r(\bX)\right)+\left(\sum_{i=1}^nW_{ni}(\bX,\Theta)-1\right) r(\bX)\right]\right]^2\\
& \quad \leq \mathbb E\left[ \sum_{i=1}^n W_{ni}(\bX,\Theta)\left( r(\bX_i)- r(\bX)\right)+\left(\sum_{i=1}^nW_{ni}(\bX,\Theta)-1\right) r(\bX)\right]^2,
\end{align*}
where, in the last step, we used Jensen's inequality. Consequently,
\begin{align}
&\mathbb E \left [ \tilde r_n(\bX)-r(\bX)\right]^2 \nonumber\\
& \quad \leq \mathbb E\left[ \sum_{i=1}^n W_{ni}(\bX,\Theta)\left(r(\bX_i)-r(\bX)\right)\right]^2+ \mathbb E \left[ r(\bX)\,\mathbf 1_{\mathcal E_n^c(\bX,\Theta)} \right]^2\nonumber\\
& \quad \leq \mathbb E\left[ \sum_{i=1}^n W_{ni}(\bX,\Theta)\left( r(\bX_i)- r(\bX)\right)\right]^2+ \left [\sup_{\mathbf x \in [0,1]^d} r^2(\bx )\right]\mathbb P\left (\mathcal E_n^c(\bX,\Theta)\right) \label{ulm1}.
\end{align}

Let us examine the first term on the right-hand side of (\ref{ulm1}). Observe that, by the Cauchy-Schwarz inequality,
\begin{align*}
&  \mathbb E \left [  \sum_{i=1}^n  W_{ni}(\bX,\Theta) \left(r(\bX_i)-r(\bX)\right)\right]^2\\
&\quad \leq  \mathbb E \left [  \sum_{i=1}^n \sqrt{W_{ni}(\bX,\Theta)}\sqrt{W_{ni}(\bX,\Theta)} \left| r(\bX_i)-r(\bX)\right|\right]^2\\
& \quad \leq \mathbb E \left [\left ( \sum_{i=1}^n  W_{ni}(\bX,\Theta) \right )\left (\sum_{i=1}^n  W_{ni}(\bX,\Theta) \left (r(\bX_i)-r(\bX)\right)^2\right)\right]\\
& \quad \leq  \mathbb E \left [\sum_{i=1}^n W_{ni}(\bX,\Theta)\left(r(\bX_i)-r(\bX)\right)^2\right]\\
& \qquad \mbox{(since the weights are subprobability weights)}.
\end{align*}
Thus, denoting by $\|\bX\|_{\mathcal S}$ the norm of $\bX$ evaluated over the components in $\mathcal S$, we obtain
\begin{align*}
&\mathbb E \left [  \sum_{i=1}^n  W_{ni}(\bX,\Theta) \left( r(\bX_i)-r(\bX)\right)\right]^2 \\
& \quad \leq \mathbb E \left [\sum_{i=1}^n W_{ni}(\bX,\Theta)\left(r^{\star}(\bX_{i\mathcal S})-r^{\star}(\bX_{\mathcal S})\right)^2\right]\\
& \quad \leq L^2  \sum_{i=1}^n \mathbb E \left [W_{ni}(\bX,\Theta)\|\bX_i-\bX\|_{\mathcal S}^2\right]\\
& \quad = nL^2 \mathbb E \left [ W_{n1}(\bX,\Theta)\|\bX_1-\bX\|_{\mathcal S}^2\right]\\
& \qquad \mbox{(by symmetry)}.
\end{align*}
But
\begin{align*}
&  \mathbb E \left [ W_{n1}(\bX,\Theta)\|\bX_1-\bX\|_{\mathcal S}^2\right]\\
 &\quad =\mathbb E \left [ \|\bX_1-\bX\|_{\mathcal S}^2 \frac{ \mathbf 1_{[\bX_1 \in A_n(\bX,\Theta)]}}{N_n(\bX,\Theta)}\,\mathbf 1_{\mathcal E_n(\bX,\Theta)}\right]\\
 & \quad =\mathbb E \left [ \|\bX_1-\bX\|_{\mathcal S}^2 \frac{ \mathbf 1_{[\bX_1 \in A_n(\bX,\Theta)]}}{1+\sum_{i=2}^n \mathbf 1_{[\bX_i \in A_n(\bX, \Theta)]}}\right]\\
 &\quad =\mathbb E \left [\mathbb E  \left [ \|\bX_1-\bX\|_{\mathcal S}^2 \frac{ \mathbf 1_{[\bX_1 \in A_n(\bX,\Theta)]}}{1+\sum_{i=2}^n \mathbf 1_{[\bX_i \in A_n(\bX, \Theta)]}}\,|\, \bX, \bX_1, \Theta\right] \right].
 \end{align*}
 Thus,
 \begin{align*}
 &  \mathbb E \left [ W_{n1}(\bX,\Theta)\|\bX_1-\bX\|_{\mathcal S}^2\right]\\
& \quad =\mathbb E \left [ \|\bX_1-\bX\|_{\mathcal S}^2\mathbf 1_{[\bX_1 \in A_n(\bX,\Theta)]}\mathbb E  \left [ \frac{ 1}{1+\sum_{i=2}^n \mathbf 1_{[\bX_i \in A_n(\bX, \Theta)]}}\,|\, \bX, \bX_1, \Theta\right] \right]\\
& \quad =\mathbb E \left [ \|\bX_1-\bX\|_{\mathcal S}^2\mathbf 1_{[\bX_1 \in A_n(\bX,\Theta)]}\mathbb E  \left [ \frac{1}{1+\sum_{i=2}^n \mathbf 1_{[\bX_i \in A_n(\bX, \Theta)]}}\,|\, \bX, \Theta\right] \right]\\
& \qquad \mbox{(by the independence of the random variables } \bX, \bX_1, \hdots, \bX_n, \Theta).
\end{align*}
By Fact \ref{FAIT2} and technical Lemma \ref{bosq},
$$
\mathbb E  \left [ \frac{1}{1+\sum_{i=2}^n \mathbf 1_{[\bX_i \in A_n(\bX, \Theta)]}}\,|\, \bX,\Theta\right] \leq \frac{2^{\lk}}{n}\leq \frac{2k_n}{n}.$$
Consequently,
\begin{align*}
&\mathbb E \left [  \sum_{i=1}^n  W_{ni}(\bX,\Theta) \left(r(\bX_i)-r(\bX)\right)\right]^2\\
& \quad \leq 2L^2k_n\mathbb E \left [ \|\bX_1-\bX\|_{\mathcal S}^2\mathbf 1_{[\bX_1 \in A_n(\bX,\Theta)]}\right].
\end{align*}
Letting
$$A_n(\bX,\Theta)=\prod_{j=1}^d A_{nj}(\bX,\Theta),$$
we obtain
\begin{align*}
& \mathbb E \left [  \sum_{i=1}^n  W_{ni}(\bX,\Theta) \left( r(\bX_i)- r(\bX)\right)\right]^2\\
& \quad \leq 2L^2k_n\sum_{j \in \mathcal S} \mathbb E  \left [ |\bX_1^{(j)}-\bX^{(j)}|^2\mathbf 1_{[\bX_1 \in A_n(\bX,\Theta)]}\right]\\
& \quad = 2L^2k_n \sum_{j \in \mathcal S} \mathbb E \left [ \rho_j(\bX,\bX_1,\Theta) \mathbb E_{\bX_1^{(j)}} \left [ |\bX_1^{(j)}-\bX^{(j)}|^2\mathbf 1_{[\bX_1^{(j)} \in A_{nj}(\bX,\Theta)]}\right]\right]
\end{align*}
where, in the last equality, we set
$$ \rho_j(\bX,\bX_1,\Theta)= \prod_{t=1, \hdots ,d, t\neq j} \mathbf 1_{[\bX_1^{(t)} \in A_{nt}(\bX,\Theta)]}.$$
Therefore, using the fact that $\bX_1$ is uniformly distributed over $[0,1]^d$,
\begin{align*}
& \mathbb E \left [  \sum_{i=1}^n  W_{ni}(\bX,\Theta) \left(r(\bX_i)-r(\bX)\right)\right]^2\\
&  \quad \leq  2L^2k_n \sum_{j \in \mathcal S} \mathbb E \left [ \rho_j(\bX,\bX_1,\Theta) \lambda^3 \left (A_{nj}(\bX,\Theta)\right) \right].
\end{align*}
Observing that 
\begin{align*}
&\lambda \left (A_{nj}(\bX,\Theta)\right) \times  \mathbb E_{[\bX_1^{(t)}\,:\,t=1, \hdots, d, t\neq j]} \left [\rho_j(\bX,\bX_1,\Theta)\right]\\
& \quad = \lambda \left ( A_n(\bX, \Theta)\right)\\
& \quad =2^{-\lk}\\
& \qquad (\mbox{Fact \ref{FAIT2}}),
\end{align*}
we are led to
\begin{align*}
&\mathbb E \left [  \sum_{i=1}^n  W_{ni}(\bX,\Theta) \left(r(\bX_i)-r(\bX)\right)\right]^2 \\
& \quad \leq 2 L^2\sum_{j \in \mathcal S} \mathbb E \left [\lambda^2 \left (A_{nj}(\bX,\Theta)\right) \right]\\
& \quad =2L^2 \sum_{j \in \mathcal S} \mathbb  EÊ\left [ 2^{-2K_{nj}(\bX,\Theta)}\right]\\
& \quad =2L^2 \sum_{j \in \mathcal S} \mathbb E \left[ \mathbb  EÊ\left [ 2^{-2K_{nj}(\bX,\Theta)}\,|\,\bX\right]\right],
\end{align*}
where, conditionally on $\bX$, $K_{nj}(\bX,\Theta)$ has a binomial $\mathcal B(\lk,p_{nj})$ distribution (Fact \ref{FAIT1}). Consequently,
\begin{align*}
& \mathbb E \left [  \sum_{i=1}^n  W_{ni}(\bX,\Theta) \left(r(\bX_i)-r(\bX)\right)\right]^2\\  
& \quad \leq  2 L^2\sum_{j \in \mathcal S}\left (1-0.75p_{nj}\right)^{\lk}\\
& \quad \leq 2 L^2\sum_{j \in \mathcal S}\exp \left ( -\frac{0.75}{\log 2} p_{nj} \log k_n\right)\\
& \quad = 2L^2 \sum_{j \in \mathcal S} \frac{1}{k_n^{\frac{0.75}{S\log 2} (1+\xi_{nj})}}\\
& \quad \leq  \frac{2 SL^2}{k_n^{\frac{0.75}{S\log 2} (1+\gamma_{n})}},
\end{align*}
with $\gamma_n=\min_{j\in \mathcal S} \xi_{nj}$.
\medskip

To finish the proof, it remains to bound the second term on the right-hand side of (\ref{ulm1}), which is easier. Just note that
\begin{align*}
 \mathbb P\left (\mathcal E_n^c(\bX,\Theta)\right)& = \mathbb  P\left (\sum_{i=1}^n \mathbf 1_{[\bX_i\in A_n(\bX,\Theta)]}=0\right)\\
 & =\mathbb E\left [ \mathbb  P\left (\sum_{i=1}^n \mathbf 1_{[\bX_i\in A_n(\bX,\Theta)]}=0\,|\,\bX,\Theta\right)\right]\\
&   = \left (1-2^{-\lk}\right)^n\\
& \quad (\mbox{by Fact \ref{FAIT2}})\\
& \leq e^{-n/2k_n}.
\end{align*}
\medskip

Putting all the pieces together, we finally conclude that
$$\mathbb E \left [ \tilde r_n(\bX)-r(\bX)\right]^2 \leq  \frac{2SL^2}{k_n^{\frac{0.75}{S\log 2} (1+\gamma_{n})}} 
+ \left[\sup_{\mathbf x \in [0,1]^d}r^2(\mathbf x)\right]e^{-n/2k_n},$$
as desired.
\subsection{Some technical results}
The following result is an extension of Lemma 4.1 in {Gy\"orfi} et al.~\cite{Laciregressionbook}. Its proof is given here for the sake of completeness.
\begin{lem}
\label{bosq}
Let $Z$ be a binomial $\mathcal B(N,p)$ random variable, with $p \in (0,1]$. Then
\begin{enumerate}
\item[$(i)$] $$\mathbb E \left [ \frac{1}{1+Z}\right]\leq \frac{1}{(N+1)p}.$$
\item[$(ii)$] $$\mathbb E \left [\frac{1}{Z}\mathbf 1_{[Z\geq 1]} \right]\leq \frac{2}{(N+1)p}.$$
\item[$(iii)$] $$\mathbb E \left [ \frac{1}{1+Z^2}\right]\leq \frac{3}{(N+1)(N+2)p^2}.$$
\end{enumerate}
\end{lem}
\noindent {\bf Proof of Lemma \ref{bosq}}\quad To prove statement $(i)$, we write
\begin{align*}
\mathbb E \left [ \frac{1}{1+Z}\right] & = \sum_{j=0}^N \frac{1}{1+j} {{N}\choose{j}}p^j(1-p)^{N-j}\\
& = \frac{1}{(N+1)p} \sum_{j=0}^N   {{N+1}\choose{j+1}}p^{j+1}(1-p)^{N-j} \\
& \leq \frac{1}{(N+1)p} \sum_{j=0}^{N+1}   {{N+1}\choose{j}}p^{j}(1-p)^{N+1-j} \\
& = \frac{1}{(N+1)p}.
\end{align*}
The second statement follows from the inequality
$$\mathbb E \left [\frac{1}{Z} \mathbf 1_{[Z\geq 1]} \right]\leq \mathbb E \left [\frac{2}{1+Z}\right]$$
and the third one by observing that
$$
\mathbb E \left [\frac{1}{1+Z^2}\right]= \sum_{j=0}^N \frac{1}{1+j^2} {{N}\choose{j}}p^j(1-p)^{N-j}.
$$
Therefore
\begin{align*}
\mathbb E \left [\frac{1}{1+Z^2}\right]&  = \frac{1}{(N+1)p}\sum_{j=0}^N \frac{1+j}{1+j^2} {{N+1}\choose{j+1}}p^{j+1}(1-p)^{N-j}\\
& \leq \frac{3}{(N+1)p}\sum_{j=0}^{N} \frac{1}{2+j} {{N+1}\choose{j+1}}p^{j+1}(1-p)^{N-j}\\
& \leq  \frac{3}{(N+1)p}\sum_{j=0}^{N+1} \frac{1}{1+j} {{N+1}\choose{j}}p^{j}(1-p)^{N+1-j}\\
& \leq \frac{3}{(N+1)(N+2)p^2}\\
& \quad (\mbox{by }(i)).
\end{align*}
\hfill $\blacksquare$
\begin{lem}
Let $Z_1$ and $Z_2$ be two independent binomial $\mathcal B(N,p)$ random variables. Set, for all $z \in \mathbb C^{\star}$, $\varphi(z)=\mathbb E [z^{Z_1-Z_2}]$. Then
\begin{enumerate}
\label{Cauchy}
\item[$(i)$] For all $z \in \mathbb C^{\star}$,
$$\varphi(z)=\left[p(1-p)(z+z^{-1})+1-2p(1-p)\right]^N.$$
\item[$(ii)$] For all $j\in \mathbb N$,
$$\mathbb P(Z_1-Z_2=j)=\frac{1}{2\pi i} \int_{\Gamma} \frac{\varphi(z)}{z^{j+1}}\emph{d}z,$$
where $\Gamma$ is the positively oriented unit circle.
\item[$(iii)$] For all $d \geq 1$,
$$\mathbb E \left [ 2^{-d(Z_1-Z_2)_+} \right]\leq \frac{24}{\pi} \int_0^1 \exp\left(-4Np(1-p)t^2\right)\emph{d}t.$$
\end{enumerate}
\end{lem}
\noindent {\bf Proof of Lemma \ref{Cauchy}}\quad Statement $(i)$ is clear and $(ii)$ is an immediate consequence of Cauchy's integral formula (Rudin \cite{Rudin}). To prove statement $(iii)$, write
\begin{align*}
\mathbb E \left [ 2^{-d(Z_1-Z_2)_+}\right] &=\sum_{j=0}^N 2^{-dj}\,\mathbb P\left ((Z_1-Z_2)_+=j\right)\\
& = \sum_{j=0}^N 2^{-dj}\,\mathbb P\left (Z_1-Z_2=j\right)\\
& \leq  \sum_{j=0}^{\infty} 2^{-dj}\,\mathbb P\left (Z_1-Z_2=j\right)\\
& = \frac{1}{2\pi i} \int_{\Gamma} \frac{\varphi (z)}{z} \sum_{j=0}^\infty \left (\frac{2^{-d}}{z}\right)^j\mbox{d}z\\
& \quad (\mbox{by statement } (ii))\\
& = \frac{1}{2\pi} \int_{-\pi}^{\pi}\frac{\varphi(e^{i\theta})}{1-2^{-d} e^{-i\theta}}\mbox{d}\theta\\
& \quad (\mbox{by setting }z=e^{i\theta}, \theta \in [-\pi,\pi])\\
& =\frac{2^{d-1}}{\pi} \int_{-\pi}^{\pi} \left[1+2p(1-p)(\cos \theta-1)\right]^N\frac{e^{i\theta}}{2^de^{i\theta}-1}\mbox{d}\theta\\
& \quad (\mbox{by statement } (i)).\\
\end{align*}
Noting that
$$\frac{e^{i\theta}}{2^de^{i\theta}-1} = \frac{2^d-e^{i\theta}}{2^{2d}-2^{d+1}\cos \theta +1},$$
we obtain
\begin{align*}
& \mathbb E \left [ 2^{-d(Z_1-Z_2)_+}\right] \\
& \quad \leq \frac{2^{d-1}}{\pi}\int_{-\pi}^{\pi}\left[1+2p(1-p)(\cos \theta-1)\right]^N \frac{2^d-\cos \theta}{2^{2d}-2^{d+1}\cos \theta +1}\mbox{d}\theta.
\end{align*}
The bound 
$$\frac{2^d-\cos \theta}{2^{2d}-2^{d+1}\cos \theta+1} \leq \frac{2^d+1}{(2^d-1)^2}$$
leads to
\begin{align*}
&\mathbb E \left [ 2^{-d(Z_1-Z_2)_+}\right] \\
&\quad \leq  \frac{2^{d-1}(2^d+1)}{\pi (2^d-1)^2}  \int_{-\pi}^{\pi} \left[1+2p(1-p)(\cos \theta-1)\right]^N\mbox{d}\theta\\
& \quad = \frac{2^{d}(2^d+1)}{\pi (2^d-1)^2}  \int_{0}^{\pi} \left[1+2p(1-p)(\cos \theta-1)\right]^N\mbox{d}\theta\\
&  \quad = \frac{2^{d}(2^d+1)}{\pi (2^d-1)^2}  \int_{0}^{\pi} \left[1-4p(1-p)\sin^2(\theta/2)\right]^N\mbox{d}\theta\\
& \qquad(\cos \theta -1=-2 \sin^2(\theta/2))\\
&\quad = \frac{2^{d+1}(2^d+1)}{\pi (2^d-1)^2}  \int_{0}^{\pi/2} \left[1-4p(1-p)\sin^2\theta\right]^N\mbox{d}\theta.
\end{align*}
Using the elementary inequality $(1-z)^N\leq e^{-Nz}$ for $z\in [0,1]$ and the change of variable
$$t=\tan(\theta/2),$$
we finally obtain
\begin{align*}
&\mathbb E \left [ 2^{-d(Z_1-Z_2)_+}\right] \\
& \quad \leq  \frac{2^{d+2}(2^d+1)}{\pi(2^d-1)^2} \int_0^1 \exp\left(-\frac{16Np(1-p)t^2}{(1+t^2)^2}\right)\frac{1}{1+t^2}\mbox{d}t\\
& \quad \leq C_d\int_0^1 \exp\left(-4Np(1-p)t^2\right)\mbox{d}t,
\end{align*}
with
$$C_d=\frac{2^{d+2}(2^d+1)}{\pi(2^d-1)^2}.$$
The conclusion follows by observing that $C_d \leq 24/\pi$ for all $d \geq 1$.
\hfill $\blacksquare$\\

Evaluating the integral in statement $(iii)$ of Lemma \ref{Cauchy} leads to the following proposition:
\begin{pro}
\label{corocauchy}
Let $Z_1$ and $Z_2$ be two independent binomial $\mathcal B(N,p)$ random variables, with $p \in (0,1)$. Then, for all $d\geq 1$, 
$$\mathbb E \left [ 2^{-d(Z_1-Z_2)_+} \right]\leq \frac{24}{\pi} \min \left (1, \sqrt{\frac{\pi}{16Np(1-p)}}\right).$$
\end{pro}
\paragraph{Acknowledgments.} I greatly thank the Action Editor and two referees for valuable comments and insightful suggestions, which lead to a substantial improvement of the paper. I would also like to thank my colleague Jean-Patrick Baudry for his precious help on the simulation section. \bibliographystyle{plain}
\bibliography{biblio}
\end{document}